%% file: neurips_2025.tex
\documentclass{article}

\PassOptionsToPackage{numbers, compress}{natbib}

\usepackage[preprint]{neurips_2025}

\usepackage[utf8]{inputenc} 
\usepackage[T1]{fontenc}    
\usepackage{hyperref}       
\usepackage{url}            
\usepackage{booktabs}       
\usepackage{amsfonts}       
\usepackage{nicefrac}       
\usepackage{microtype}      
\usepackage[table, dvipsnames]{xcolor}         
\usepackage{xpatch}
\usepackage{tabularx}
\usepackage{longtable}
\usepackage{arydshln} 
\usepackage{graphicx} 
\usepackage{wrapfig}
\usepackage{subcaption}
\usepackage{amssymb}
\usepackage{mathtools}
\usepackage{amsthm}
\usepackage{amsmath}
\usepackage{bbding}
\usepackage{multirow}
\usepackage{makecell}
\usepackage{booktabs}
\usepackage{enumitem}
\usepackage{colortbl}
\usepackage{wrapfig}
\usepackage{todonotes}
\usepackage{fontawesome}

\usepackage{xpatch}
\makeatletter
\xapptocmd{\NAT@bibsetnum}{\setlength{\leftmargin}{0pt}\setlength{\itemindent}{\labelwidth}\addtolength{\itemindent}{\labelsep}}{}{}
\makeatother

\include{Mask_QA/math}

\newcommand{\think}[1]{%
    \textcolor{RoyalBlue}{\texttt{<think>}} 
    #1 
    \textcolor{RoyalBlue}{\texttt{</think>}}
}

\newcommand{\search}[1]{%
    \textcolor{YellowOrange}{\texttt{<search>}} 
    #1 
    \textcolor{YellowOrange}{\texttt{</search>}} 
}

\newcommand{\information}[1]{%
    \textcolor{Periwinkle}{\texttt{<information>}} 
    #1 
    \textcolor{Periwinkle}{\texttt{</information>}} 
}

\newcommand{\answer}[1]{%
    \textcolor{OliveGreen}{\texttt{<answer>}} 
    #1 
    \textcolor{OliveGreen}{\texttt{</answer>}} 
}

\newcommand{\masksearch}{\textsc{MaskSearch}}
\usepackage{multicol}
\newcommand{\symboletongyi}{\raisebox{0pt}{~\includegraphics[scale=0.012]{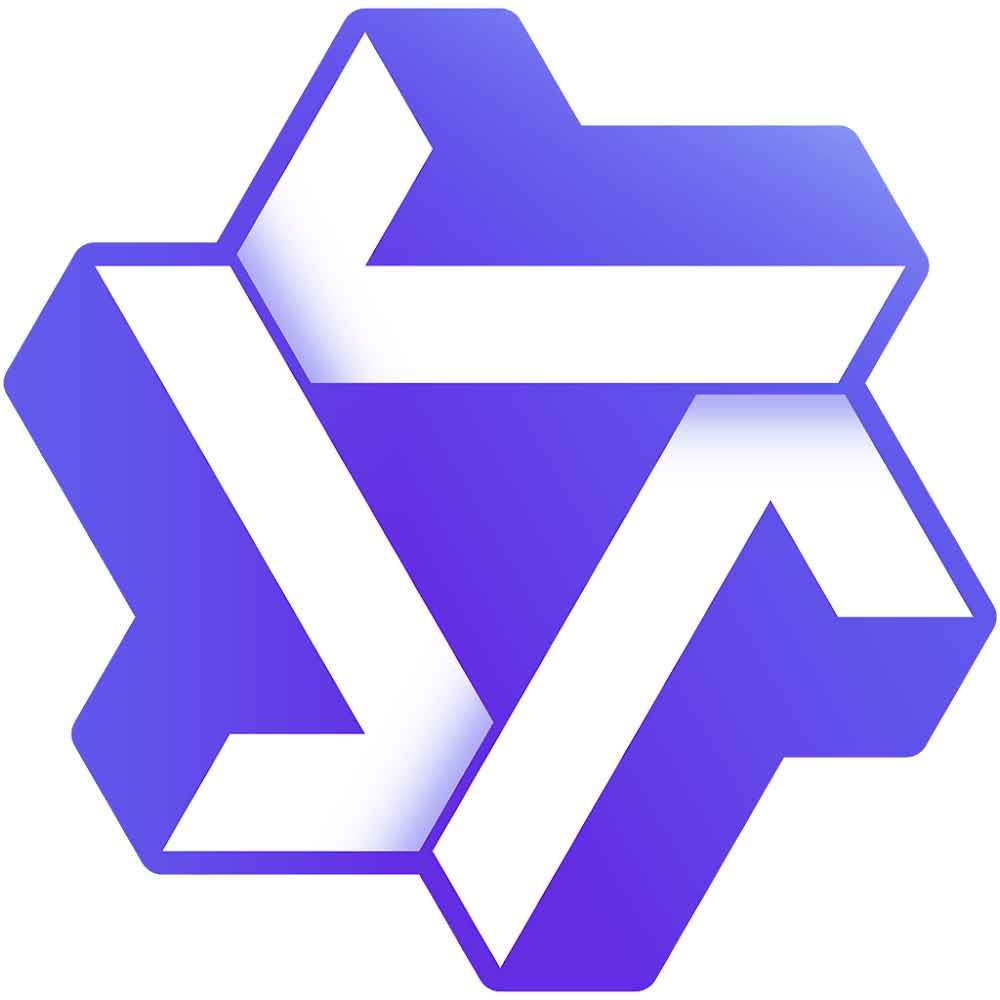}}~}

\title{\masksearch: A Universal Pre-Training Framework \\to Enhance Agentic Search Capability}

\makeatletter
\def\@fnsymbol#1{\ensuremath{\ifcase#1\or \dagger\or *\or \ddagger\or
   \mathsection\or \mathparagraph\or \|\or **\or \dagger\dagger
   \or \ddagger\ddagger \else\@ctrerr\fi}}
\makeatother

\author{%
Weiqi Wu$^{\dagger}$, Xin Guan\thanks{Equal Contribution. This work was done during Weiqi Wu and Xin Guan’s internship at Tongyi Lab, Alibaba Group.}, Shen Huang, Yong Jiang$^{*}$, Pengjun Xie, \\
\textbf{Fei Huang, Jiuxin Cao, Hai Zhao\thanks{Corresponding authors.}, Jingren Zhou}\\
Tongyi Lab\symboletongyi, Alibaba Group \\
}

\begin{document}

\maketitle

\begin{abstract}
    Retrieval-Augmented Language Models (RALMs) represent a classic paradigm where models enhance generative capabilities using external knowledge retrieved via a specialized module. Recent advancements in Agent techniques enable Large Language Models (LLMs) to autonomously utilize tools for retrieval, planning, and reasoning. 
    While existing training-based methods show promise, their agentic abilities are limited by inherent characteristics of the task-specific data used during training.
    To further enhance the universal search capability of agents, we propose a novel pre-training framework, \textbf{\masksearch}. In the pre-training stage, we introduce the Retrieval Augmented Mask Prediction (RAMP) task, where the model learns to leverage search tools to fill masked spans on a large number of pre-training data, thus acquiring universal retrieval and reasoning capabilities for LLMs. After that, the model is trained on downstream tasks to achieve further improvement. 
    We apply both Supervised Fine-tuning (SFT) and Reinforcement Learning (RL) for training. For SFT, we combine agent-based and distillation-based methods to generate training data, starting with a multi-agent system consisting of a planner, rewriter, observer, and followed by a self-evolving teacher model. While for RL, we employ DAPO as the training framework and adopt a hybrid reward system consisting of answer rewards and format rewards. Additionally, we introduce a curriculum learning approach that allows the model to learn progressively from easier to more challenging instances based on the number of masked spans. We evaluate the effectiveness of our framework in the scenario of open-domain multi-hop question answering. Through extensive experiments, we demonstrate that \textsc{MaskSearch} significantly enhances the performance of LLM-based search agents on both in-domain and out-of-domain downstream tasks. Code is available at \url{https://github.com/Alibaba-NLP/MaskSearch}.
\end{abstract}

\input{Mask_QA/section/introduction}

\input{Mask_QA/section/related_works}
\input{Mask_QA/section/method}

\input{Mask_QA/section/experiment_setup}
\input{Mask_QA/section/exp_results}

\input{Mask_QA/section/discussion}
\input{Mask_QA/section/conclusion}

\newpage

\bibliographystyle{unsrtnat}  
\small
\bibliography{Reference}
\normalsize


\newpage
\appendix

\input{Mask_QA/appendix/training_setup} \label{sec: setup}
\input{Mask_QA/appendix/ralm}
\input{Mask_QA/appendix/results} \label{sec:exp_results}
\section{Prompt Demonstration}
\input{Mask_QA/appendix/prompt}

\section{Reinforcement Learning} \label{appendix:rl}
\input{Mask_QA/appendix/RL_appendix}

\input{Mask_QA/appendix/case}

\input{Mask_QA/appendix/Impacts_Ethics}


\end{document}

%% file: Mask_QA/math.tex
\def\sD{{\mathbb{D}}}

%% file: Mask_QA/section/introduction.tex
\section{Introduction}

Large Language Models (LLMs) \cite{qwen2025qwen25technicalreport, DeepSeekAI2025DeepSeekR1IR, grattafiori2024llama3herdmodels, Minaee2024LargeLM} demonstrate strong performance across a variety of tasks by leveraging vast internal knowledge \cite{chang2024largelanguagemodelsacquire, wu-etal-2023-plms, Petroni2019LanguageMA}, but suffer from hallucinations and often fall short in effectively addressing domain-specific or real-time tasks \cite{Ding2024ASO, Lewis2020RetrievalAugmentedGF}. Retrieval-Augmented Language Models (RALMs) \cite{ram2023incontextretrievalaugmentedlanguagemodels, Lewis2020RetrievalAugmentedGF, guu2020realm} have been proposed to enhance LLMs by incorporating external knowledge, where a retrieval mechanism fetches information relevant to the input to augment the model's generation. While this approach has been successful, the separation of retrieval and generation limits the model's adaptability, preventing it from proactively acquiring information needed for multi-step tasks.

With autonomous AI agents gaining momentum, a new paradigm has emerged—agents leveraging search engines as tools while employing agentic strategies such as planning, reasoning, reflection, and multi-agent collaboration to enhance their problem-solving capabilities. While prompt-based workflows have been widely used \cite{li2025searcho1agenticsearchenhancedlarge}, they suffer from inefficiency and a lack of flexibility. Training an LLM-based search agent offers a more promising solution \cite{jin2025searchr1trainingllmsreason}, yet current methods primarily rely on task-specific data, which limits the ability to generalize across a broader range of tasks.

\begin{figure}[t]
    \centering
    \includegraphics[width=0.95\linewidth]{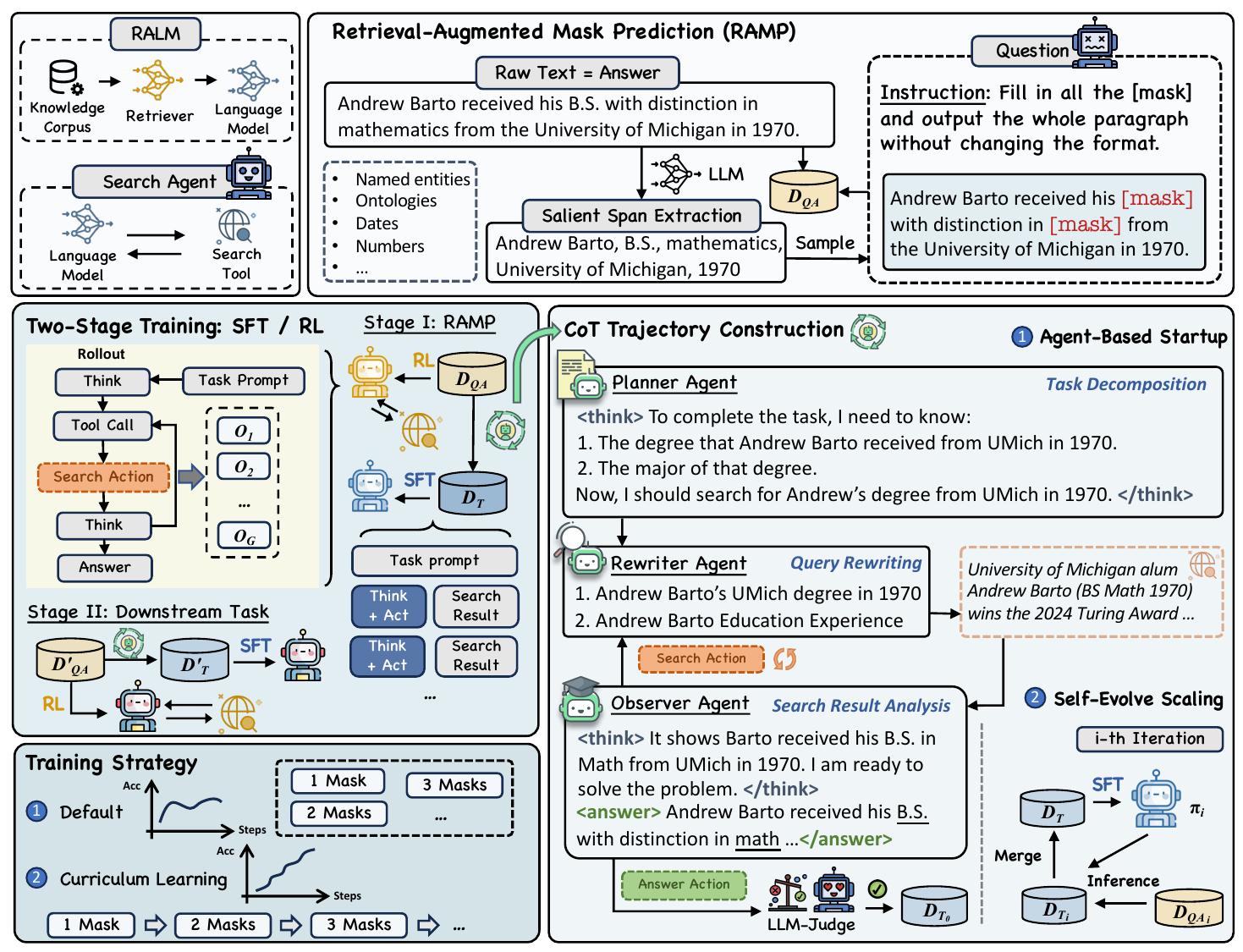}
    \caption{Overview of \masksearch, a pre-training framework to incentivize the agentic RAG capabilities of LLMs. Based on the Retrieval-Augmented Mask Prediction (RAMP) task, models can be trained via SFT or RL to acquire generalizable abilities before downstream task training.}
    \label{fig:overview}
\vspace{-10pt}
\end{figure}

Previous research on Masked Language Models (MLM) \cite{guu2020realm, devlin-etal-2019-bert} has proved that models' memory, understanding and generation capabilities can be effectively incentivized through nearly infinite, diverse and verifiable data. By allowing models to think and utilize search tools to fill in the masks instead of solely based on the given context, we can further enhance their tool usage and reasoning abilities.
Building on these insights, we propose \textbf{\masksearch}, a universal pre-training framework designed to enhance the agentic search capabilities of LLMs within a unified model architecture. 
The model is trained on a theoretically infinitely scalable task named Retrieval-Augmented Mask Prediction (RAMP), where it learns to fill in masked spans by performing multi-step search and reasoning as illustrated in Figure \ref{fig:overview}. The task requires general capabilities of task decomposition, search tool utilization and observation-based reasoning, which are highly transferable to downstream tasks such as open-domain question answering.

Both Supervised Fine-tuning (SFT) and Reinforcement Learning (RL) can be used for training on RAMP. To construct the SFT dataset, we propose a method that combines agent-based and distillation-based approaches to generate Chain-of-Thought (CoT) trajectories that can solve the RAMP task. Initially, a multi-agent system involving a planner, rewriter, and observer agent is utilized to synthesize reasoning data. Subsequently, we adopt an iterative self-evolutionary distillation strategy, using an increasingly fine-tuned model to generate the next partition of the dataset.
For RL, we employ the Dynamic Sampling Policy Optimization (DAPO) algorithm \cite{yu2025dapoopensourcellmreinforcement} to optimize the model's search and reasoning process with a hybrid reward system including rule-based format rewards and model-based answer rewards. In addition, we introduce a curriculum learning method based on the number of masks, enabling the model to learn progressively from easier to more difficult scenarios.

Extensive analysis shows that incorporating RAMP as the pre-training task yields significant performance enhancements across a variety of open-domain question-answering datasets. It not only provides a stable improvement in recall scores for the in-domain downstream task but also achieves pronounced gains on out-of-domain benchmarks. We also validate the scalability of \masksearch{} by constructing a 10M pertaining dataset for SFT. Furthermore, despite the fact that RL is not typically used for pre-training, our experiments indicate that it achieves impressive performance gains through pre-training. It even demonstrates a greater potential to optimize model performance, achieving higher upper limits after post-training compared to using SFT. These findings highlight the effectiveness of \masksearch{} in enhancing the general agentic search capabilities of LLMs, regardless of their size, type, or the specific training method employed.

%% file: Mask_QA/section/related_works.tex
\section{Related Work}

\paragraph{Retrieval-Augmented Generation}
RAG is a powerful approach to bridging the gap between static model parameters and dynamic external knowledge \cite{Ding2024ASO, Yan2024CorrectiveRA, Gao2023RetrievalAugmentedGF, Lewis2020RetrievalAugmentedGF}. By incorporating an external retrieval module, LLMs can generate responses with real-time or domain-specific \cite{Zhang2024RAFTAL, Gilbert2024AugmentedNL} data, thereby mitigating issues like hallucinations \cite{li2024enhancingllmfactualaccuracy, Huang2023ASO} and outdated facts. Early advances demonstrate that coupling neural retrievers with language models improves factual accuracy in tasks such as open-domain question answering \cite{Lewis2020RetrievalAugmentedGF, guu2020realm}. Recent studies have advanced RAG by embedding autonomous AI agents into the RAG workflow \cite{singh2025agenticretrievalaugmentedgenerationsurvey, ravuru2024agenticretrievalaugmentedgenerationtime, an2024goldenretrieverhighfidelityagenticretrieval}, leveraging the advantage of agent frameworks like planning, reflection and multi-agent collaboration \cite{wang2025vidoragvisualdocumentretrievalaugmented}.

\paragraph{Agent} 
The advent of LLM-based agents marks a revolutionary transformation in the AI domain. These agents can autonomously plan, reason, utilize tools, and retain memory while interacting with dynamic environments, thereby being capable of handling complex tasks such as web search \cite{he2025pasallmagentcomprehensive, alzubi2025opendeepsearchdemocratizing} and operations \cite{Gur2023ARW}, creative writing \cite{zhang2024codeagent}, and travel planning \cite{gundawar2024robustplanningllmmoduloframework}. In addition to realizing LLM agents through prompt engineering \cite{shinn2024reflexion, park2023generativeagentsinteractivesimulacra, yao2023react, AutoAgent}, recent research has also focused on optimizing and training these agent skills in an end-to-end manner \cite{putta2024agent, feng2024agile}. We aim to integrate this idea into the training of RALM, enabling RALM to more autonomously retrieve knowledge and enhance its understanding, reasoning and generation capabilities.

\paragraph{Reasoning LLM}

Improving reasoning capabilities in LLMs has become a central focus to push the frontier of coherence and consistency in complex problem-solving. The advent of Large Reasoning Models (LRMs) \cite{li2025startselftaughtreasonertools, DeepSeekAI2025DeepSeekR1IR, openai2024openaio1card} demonstrates the effectiveness of inference-time scaling \cite{muennighoff2025s1simpletesttimescaling, Snell2024ScalingLT} and multi-step reasoning \cite{Wu2023ChainOT, yao2023treethoughtsdeliberateproblem, yao2023react} mechanisms. Hence, smaller models can also tackle complex tasks in a series of logical steps rather than attempting to provide solutions in a single step. Apart from well-designed prompt-based methods \cite{li2025searcho1agenticsearchenhancedlarge, yugeswardeenoo2024questionanalysispromptingimprovesllm, fu2023complexitybasedpromptingmultistepreasoning} and Supervised Fine-tuning (SFT) \cite{subramaniam2025multiagentfinetuningselfimprovement, lin2024raditretrievalaugmenteddualinstruction, mukherjee2023orcaprogressivelearningcomplex}, where even small models show competitive performance on multi-step reasoning \cite{srivastava2025reasoningabilitysmalllanguage, shridhar-etal-2023-distilling, fu2023specializingsmallerlanguagemodels}, Reinforcement Learning (RL) has been proved to be effective for models to gain more advanced reasoning capabilities \cite{yu2025dapoopensourcellmreinforcement, jin2025searchr1trainingllmsreason, DeepSeekAI2025DeepSeekR1IR}, which has now become a mainstream training method.

%% file: Mask_QA/section/method.tex
\section{\masksearch{}}

\subsection{Preliminary}

We define Retrieval-Augmented Mask Prediction (RAMP) as the pre-training task of \masksearch{}. It involves predicting the masked spans in an input context sequence \( x \) that contains \( n \) masked spans, by proactively retrieving relevant information from an external knowledge corpus \( D \) using a search tool $\mathcal{R}$. To train an LLM-based search agent \(\pi_\theta\), which is parameterized by $\theta$ and initialized from a pre-trained base model to take a strong starting point, there are two primary methods: Supervised Fine-tuning (SFT) and Reinforcement Learning (RL).

\paragraph{Supervised Fine-tuning (SFT)}
Let \(\pi_t\) be an advanced implementation capable of multi-step agentic RAG and reasoning. The goal of SFT is to train \(\pi_\theta\) on a dataset \(\sD_t\), where each instance consists of Chain-of-Thought (CoT) data \( y_t = \pi_t(x) \). Here, \( y_t \) represents the reasoning trace generated by \(\pi_t\) to solve the RAMP task through searching and reasoning. During training, we treat the search results as a latent variable by masking the retrieved tokens and focusing on optimizing the model response.
The optimization process can be formulated as:
\begin{equation}
\pi_\theta \leftarrow \arg\min_{\pi} \mathbb{E}_{(x, y_t) \sim \sD_t} \left[ \mathcal{L}(\pi_\theta(x, D), y_t) \right]
\end{equation}

\paragraph{Reinforcement Learning (RL)} 
We follow the paradigm of Search R1 \cite{jin2025searchr1trainingllmsreason}, which incorporates the search engine $\mathcal{R}$ into the RL process for optimization. The optimization objective is formulated as: 
\begin{equation}
    \max_{\pi_\theta} \mathbb{E}_{x \sim \mathcal{D}, y \sim \pi_{\theta}(\cdot \mid x; \mathcal{R})} 
\left[ r_{\phi}(x, y) \right] 
- \beta \mathbb{D}_{\text{KL}} \left[ \pi_{\theta}(y \mid x; \mathcal{R}) \,||\, \pi_{\text{ref}}(y \mid x; \mathcal{R}) \right],
\end{equation}
 where $\pi_{\theta}$ is the policy model, $\pi_{\text{ref}}$ is the reference model, $r_{\phi}$ is the reward function and $\mathbb{D}_{\text{KL}}$ is KL-divergence measure. We employ the Dynamic Sampling Policy Optimization (DAPO) RL algorithms \cite{yu2025dapoopensourcellmreinforcement}, and mask the retrieved tokens from the search engine during gradient computation as well. Details can be found in Appendix \ref{appendix:rl}.

\subsection{Retrieval-Augmented Mask Prediction (RAMP)}

\paragraph{Question-Answer Generation} We leverage Wikipedia as our data source to build RAMP to ensure comprehensive and diverse domain coverage. Since the answers naturally originate from the unmasked portions of the original paragraphs, our task is to design a method for selectively masking spans in the paragraphs, with a focus on spans that require deeper reasoning and knowledge retrieval, rather than simple local context understanding.

\paragraph{Salient Span Extraction}

Salient Span Masking~\cite{cole2023salientspanmaskingtemporal, guu2020realm} is a critical strategy for creating challenging tasks, where the masked spans require world knowledge to predict. instead of solely local context. We broaden the definition of salient spans to include not only named entities and dates, but also ontologies, specific terms, and numerical values. 

Qwen-Turbo~\citep{qwen2025qwen25technicalreport} is utilized to extract salient spans. After extracting the spans, we randomly select $k$ spans (where $0 < k < 5$) and replace them with a mask token [mask]. This ensures the CoT trajectories remain manageable and focused while still offering sufficient challenge. The model then predicts the original content of the masked spans based on the context $x$.

\subsection{CoT Trajactory Construction}

Supervised Fine-tuning (SFT) on CoTs is the most direct way to enable multi-step reasoning. This involves generating reasoning traces that guide the model through solving the RAMP task. There are typically two ways to synthesize data: (1) agent-based, prompting a model to complete the task, and (2) distillation-based, using a stronger model to generate traces directly. We propose a hybrid approach for constructing CoT data that combines the two approaches, and construct a 10M CoT dataset (14B tokens) to validate the scalability of \masksearch{} as a pre-training framework.

\paragraph{Agent-Based Startup}
\label{sec:multi-agent}

Initially, we orchestrate a synthesizing process involving planning, search, and reflection, supported by a multi-agent system, as shown in Figure \ref{fig:overview}. The \textbf{Planner Agent} first analyzes the overall task and breaks it into sub-tasks, generating an initial search query. The \textbf{Rewriter Agent} refines the generated query for improved knowledge retrieval and calls the search tool. The \textbf{Observer Agent} reviews the search results and steps taken, determining whether the task can be resolved or if additional steps are needed, updating the process until the final answer is obtained. Each agent is instructed with a few-shot prompt, as detailed in Appendix \ref{sec: ma_prompt}. The final answer is evaluated by LLM-as-Judge and only trajectories that correctly fill all the masks are curated in \(\sD_0\). 

\paragraph{Self-Evolve Distillation}

 To rapidly scale up the dataset while maintaining high data quality, we utilize an iterative generation strategy by using a trained teacher model $\pi_{t}$ instead of the multi-agent method. During the \(j\)-th iteration, the current dataset \(\sD_j\) is used to fine-tune the model \(\pi_{\theta_{j-1}}\), resulting in an updated policy \(\pi_{\theta_j}\). This updated policy incorporates the reasoning traces and search strategies learned from \(\sD_j\). The updated policy \(\pi_{\theta_j}\) is then employed as the new teacher model \(\pi_{t_{j+1}}\), which is used to synthesize the next iteration of the dataset \(\sD_{j+1}\).
\begin{equation}
\pi_{\theta_j} \leftarrow \arg\min_{\pi} \mathbb{E}_{(x, y_t) \sim \sD_j} \left[ \mathcal{L}(\pi_{\theta_j}(x, D), y_t) \right], \quad \pi_{t_{j+1}} \leftarrow \pi_{\theta_j}
\end{equation}
\begin{equation}
\sD_{j+1} \leftarrow \{(x, y_t) \mid y_t = \pi_{t_{j+1}}(x, D)\}, \quad \sD_t \leftarrow \{\sD_0, \dots, \sD_{j+1}\}
\end{equation}
This iterative approach ensures that the model continuously learns from increasingly complex and diverse reasoning traces, as each iteration builds on the improved capabilities of the previous one.


\subsection{RL Reward Design} \label{sec:reward}
Reward signals are essential in RL, guiding the direction of model optimization. Our hybrid reward system consists of two components: a format reward and an answer reward, each contributing 50\% to a total score of 1: 
\begin{equation}
    r_{\phi}(x, y) = 0.5R_{f} (y) + 0.5R_{a} (\hat{y},y_{pre})
\end{equation}
where \(\hat{y}\) is the ground truth answer, \(y\) is the response and \(y_{pre}\) is the extracted final answer from \(y\).
The format reward \(R_{f}\) evaluates the response of the policy model to ensure it conforms to the specified answer format. It uses string matching to check whether the response contains special symbols that distinguish the various inference stages. If the response meets the format criteria, it is awarded 1 point; otherwise, it receives 0. 

For the answer reward \(R_{a}\), we explore various reward functions to identify the optimal one for optimization:

1)\textbf{ Token-level Recall Reward}: We set token-level Recall $TR$ as the reward metric. 

2) \textbf{Token-level Recall with Answer Length Penalty Reward} (Penalty-based Reward): Recall Reward often lead to reward hacking. To address this, we introduce the penalty for the answer length, defined by the following: 
\begin{equation}
R_{a} (\hat{y},y_{pre}) = TR(\hat{y},y_{pre}) - \alpha(\min(\max(\log_2 \left( \frac{|y_{pre}|}{\beta \times {|\hat{y}|}} \right), 0), \gamma))
\end{equation}
where \(\alpha\), \(\beta\), \(\gamma\) are parameters that adjust the strictness of the length penalty, set to 0.2, 8, and 4.

3) \textbf{Model-Based Reward}: We utilize the Qwen2.5-72B-Instruct model as a judge, evaluating the consistency between generated answers and standard answers, assigning a score of 0 or 1.

\subsection{Curriculum Learning}

Curriculum learning is a training strategy that sorts training samples by difficulty and presents them to the model in an incremental manner. In the context of our RAMP task, the number of masked spans $k$ serves as the primary metric for difficulty. Instead of random sampling, the curriculum learning method starts with simpler tasks containing fewer masked spans and progressively introduces more complex tasks with a higher number of masked spans. This approach allows the model to first learn fundamental reasoning skills and gradually build up its capabilities to handle more challenging scenarios. By following this curriculum, the model can better adapt to the increasing complexity of the tasks, leading to improved performance and more robust reasoning abilities.
\begin{equation}
\pi_{\theta} \leftarrow \arg\min_{\pi} \mathbb{E}_{(x, y_t) \sim \sD_{t_k}} \left[ \mathcal{L}(\pi_\theta(x, D), y_t) \right], \quad \text{for } k \text{ in } \{1, 2, 3, 4\}
\end{equation}

%% file: Mask_QA/section/experiment_setup.tex
\section{Experiment Setup}

\paragraph{Models}

\begin{wraptable}[11]{r}{5.5cm}
\setlength{\tabcolsep}{1pt}
\vspace{-1.3em}
\caption{Test datasets used in the experiment. Datasets marked with $\dag$ use their development split as the test set.}
\vspace{-0.2em}
\centering
    \small
    \resizebox{5.5cm}{!}{\begin{tabular}{lcc}
    \toprule
        \textbf{Dataset}   & \textbf{Hops} &  \textbf{\#Test}\\
    \midrule
      HotpotQA\dag \cite{yang2018hotpotqa}  &  2 & 7405\\
      \hdashline
      FanoutQA\dag \cite{zhu-etal-2024-fanoutqa} &  avg. 7 & 310\\
      Musique\dag \cite{trivedi2021musique}   &  2-4 & 2417 \\
      2WikiMultiHopQA \dag\cite{xanh2020_2wikimultihop} &  2-4  & 12576\\
      Bamboogle \cite{press-etal-2023-measuring} &  2 & 125 \\
      FreshQA \cite{vu2023freshllmsrefreshinglargelanguage} & 1-2 & 374 \\
    \bottomrule
    \end{tabular}}
    \label{tab:dataset}
\end{wraptable}
\setlength{\tabcolsep}{6pt}

We conduct experiments with two series of foundation models: (1) \textbf{\textsc{Qwen2.5}} \cite{qwen2025qwen25technicalreport}: \textsc{Qwen2.5-1.5B}, \textsc{Qwen2.5-3B} and \textsc{Qwen2.5-7B}; (2) \textbf{\textsc{LLaMa3}} \cite{grattafiori2024llama3herdmodels}: \textsc{LLaMa-3.2-1B}, {LLaMa-3.2-3B} and \textsc{LLaMa-3.1-8B}.
To generate CoT trajectories for Supervised Fine-tuning, we use \textsc{Qwen-Max} to build the multi-agent system as well as filter the correct trajectories. For self-evolve distillation, we fine-tune with \textsc{Qwen2.5-7B} when the size of the curated dataset reaches 250K, 500K and 1M.  We use the instruct models for RL, as base models often fail to follow the instructions. The detailed training setups can be found in Appendix \ref{sec: setup}.

\paragraph{Datasets}

In our study, we employ a variety of datasets to evaluate the performance of our proposed training task, as presented in Table \ref{tab:dataset}. For the downstream task, we select HotpotQA \cite{yang2018hotpotqa} as a representative challenge. During the SFT phase, we utilize the agent-based method to synthesize and filter 58K correct CoT trajectories for training data. For out-of-domain testing, we first evaluate the model on a range of multi-hop question-answering tasks, including FanoutQA \cite{zhu-etal-2024-fanoutqa}, Musique \cite{trivedi2021musique} and 2WikiMultiHopQA \cite{xanh2020_2wikimultihop}. Furthermore, we use the data without false premises in FreshQA  \cite{vu2023freshllmsrefreshinglargelanguage} to evaluate its performance on single-hop reasoning questions. 
We report token-level Recall as the evaluation metric. Specifically, it segments the generated text and the golden text into token lists and calculates the ratio of common tokens between model-generated responses and ground truth.

\paragraph{Baselines}

We compare our method against the following baselines with retrieval:
(1) \textbf{RAG-PE}: The model generates a response based on retrieval results; (2) \textbf{Agent-PE}: As elaborated in Section \ref{sec:multi-agent}, it leverages advanced prompt-based techniques to handle complex reasoning tasks; (3) \textbf{Distilled Search-R1}: The model is directly fine-tuned on the downstream data, i.e., 58K CoT trajectories from HotpotQA; (4) \textbf{Search R1} \cite{jin2025searchr1trainingllmsreason}: The model is directly 
 trained using reinforcement learning on the downstream data from HotpotQA until its performance on the validation set converges.

%% file: Mask_QA/section/exp_results.tex
\section{Experimental Results}

\begin{table}[t]
    \centering
    \caption{Evaluation results of different methods on various open-domain question answering datasets. Bold and underlined indicate the best and the second best results.}
    \label{tab:main_res}
    \resizebox{1.0\textwidth}{!}{
    \begin{tabular}{lccccccccc}
    \toprule[0.05pt]
    \toprule[0.05pt]
        \textbf{Methods} &  \textbf{Pre-Training} &  \textbf{Post-Training} & \textbf{HotpotQA} & \textbf{FanoutQA} & \textbf{Musique} & \textbf{2Wiki} & \textbf{Bamboogle} & \textbf{FreshQA} & \textbf{Avg.} \\
    \midrule
        \rowcolor{gray!20} 
        \multicolumn{10}{c}{\textit{Qwen2.5-1.5B}} \\
        RAG-PE & \XSolidBrush & \XSolidBrush & 29.45 & 27.37 & 12.07 & 37.32 & 22.93 & 41.19 & 28.37 \\
        Agent-PE & \XSolidBrush & \XSolidBrush & 48.74 & 36.15 & 28.87 & 49.72 & 58.87 & 63.27 & 47.60 \\
        Distilled Search-R1 & \XSolidBrush & SFT & 64.13 & 47.76 & 35.02 & 76.45 & 63.87 & 68.12 & 59.22 \\
        Search-R1  & \XSolidBrush & RL & 61.72 & 42.68 & 35.63 & 64.40 & 64.59 & 74.44 & 57.24 \\
         \midrule
        \multirow{4}*{\masksearch} & SFT & SFT & \underline{67.58} & \textbf{53.18} & 38.58 & \textbf{81.13} & \textbf{75.65} & 75.47 & \underline{65.27} \\
         & RL & SFT  & 66.23 & 45.61 & \underline{40.55} & 71.77 & \underline{73.79} & \underline{75.91} & 62.31 \\
         & SFT & RL  & 65.95 & 48.64 & 40.36 & 72.00 & 72.67 & 73.54 & 62.19 \\
         & RL & RL  & \textbf{71.02} & \underline{49.73} & \textbf{40.59} & \underline{76.03} & 73.60 & \textbf{81.08} & \textbf{65.34} \\
    \midrule
    \rowcolor{gray!20} 
        \multicolumn{10}{c}{\textit{Qwen2.5-3B}} \\
        RAG-PE & \XSolidBrush & \XSolidBrush & 38.37 & 41.48 & 20.78 & 51.14 & 37.60 & 61.55 & 41.82 \\
        Agent-PE & \XSolidBrush & \XSolidBrush  & 51.17 & 49.82 & 25.27 & 58.14 & 56.40 & 67.80 & 48.10 \\
        Distilled Search-R1 & \XSolidBrush & SFT & 67.38 & 54.00 & 38.20 & 79.76 & 68.05 & 77.59 & 64.17 \\
        Search-R1 & \XSolidBrush & RL & 69.03 & 48.55 & 39.08 & 78.85 & 72.53 & 76.78 & 64.14 \\
         \midrule
        \multirow{4}*{\masksearch} & SFT & SFT & \underline{69.30} & \textbf{56.03} & \underline{40.12} & \textbf{82.36} & \underline{74.52} & \underline{79.84} & \underline{67.03} \\
         & RL & SFT  & 68.23 & \underline{55.85} & 39.48 & \underline{81.72} & 73.87 & 77.58 & 66.12\\
         & SFT & RL  & 69.03 & 54.55 & 39.08 & 79.85 & 72.53 & 76.78 & 65.30 \\
         & RL & RL  & \textbf{73.08} & 53.02 & \textbf{44.48} & 80.43 & \textbf{80.13} & \textbf{85.07} & \textbf{69.37} \\
    \midrule
    \rowcolor{gray!20} 
        \multicolumn{10}{c}{\textit{Qwen2.5-7B}} \\
        RAG-PE & \XSolidBrush & \XSolidBrush & 43.55 & 51.92 & 25.05 & 53.86 & 44.60 & 64.40 & 47.23 \\
        Agent-PE & \XSolidBrush & \XSolidBrush & 61.75 & 55.69 & 34.25 & 68.77 & 63.25 & 75.81 & 58.25 \\
        Distilled Search-R1 & \XSolidBrush & SFT & 69.55 & 57.24 & 41.06 & 83.84 & 73.07 & 78.97 & 67.29 \\
        Search-R1 & \XSolidBrush & RL & 70.59 & 56.25 & 41.29 & 80.50 & 79.33 & 78.46 & 67.74 \\
         \midrule
        \multirow{4}*{\masksearch} & SFT & SFT & 70.44 & \textbf{60.85} & 41.76 & \textbf{84.65} & 80.13 & \textbf{81.12} & \underline{69.83} \\
        & RL & SFT  & 70.84 & 56.29 & 41.90 &  \underline{83.38} & 78.53 & 78.93 & 68.31 \\
        & SFT & RL  & \underline{71.69} & 57.69 & \underline{42.23} & 81.25 & \underline{81.87} & 75.42 & 68.36 \\
        & RL & RL  & \textbf{75.61} & \underline{58.96} & \textbf{45.54} & 82.10 & \textbf{83.00} & \underline{80.85} & \textbf{71.01} \\
    \bottomrule[0.05pt]
    \bottomrule[0.05pt]
    \end{tabular}
    }
    \label{tab:performance}
\end{table}

\subsection{Main Results}

\paragraph{First-stage training of the RAMP task benefits LLM-based search agents.}
As shown in Table \ref{tab:main_res}, our proposed \masksearch{}, which incorporates RAMP as the pre-training stage, significantly enhances the model performance across multiple open-domain question-answering datasets. In the in-domain dataset, i.e., HotpotQA, RAMP provides a stable improvement in the model's recall score. On out-of-domain datasets, the improvement is even more pronounced. For instance, on the Bamboogle dataset, the Qwen2.15-1.5B model achieves a substantial increase of \textbf{11.78} compared to post-training only, while the LLaMA model sees an impressive gain of \textbf{15.12} \footnote{LLaMA results in Appendix \ref{sec:exp_results}.}. 
Moreover, smaller models (e.g., Qwen2.5-1.5B) can perform comparably to larger models on a variety of tasks. This proves that RAMP, as a scalable learning signal, can help models better strengthen their abilities to decompose problems and interact with search tools.

\paragraph{RL offers higher performance gains on RAMP compared with SFT.}
While SFT proves to be effective in improving search agent performance, RL demonstrates the potential to achieve even higher upper limits when applied to RAMP tasks. The combination of RL with RAMP shows better gains over SFT alone, especially in the in-domain dataset HotpotQA, where an improvement of 3 to 5 points is achieved on different-scale models, suggesting that RL can better optimize the model for the specific nuances and challenges presented by the retrieval-augmented prompting process. This finding highlights the importance of exploring advanced training techniques like RL to fully exploit the benefits of RAMP in enhancing RALM performance. 

\subsection{Scaling Performance}

\begin{figure}[t]
    \centering
    \includegraphics[width=\linewidth]{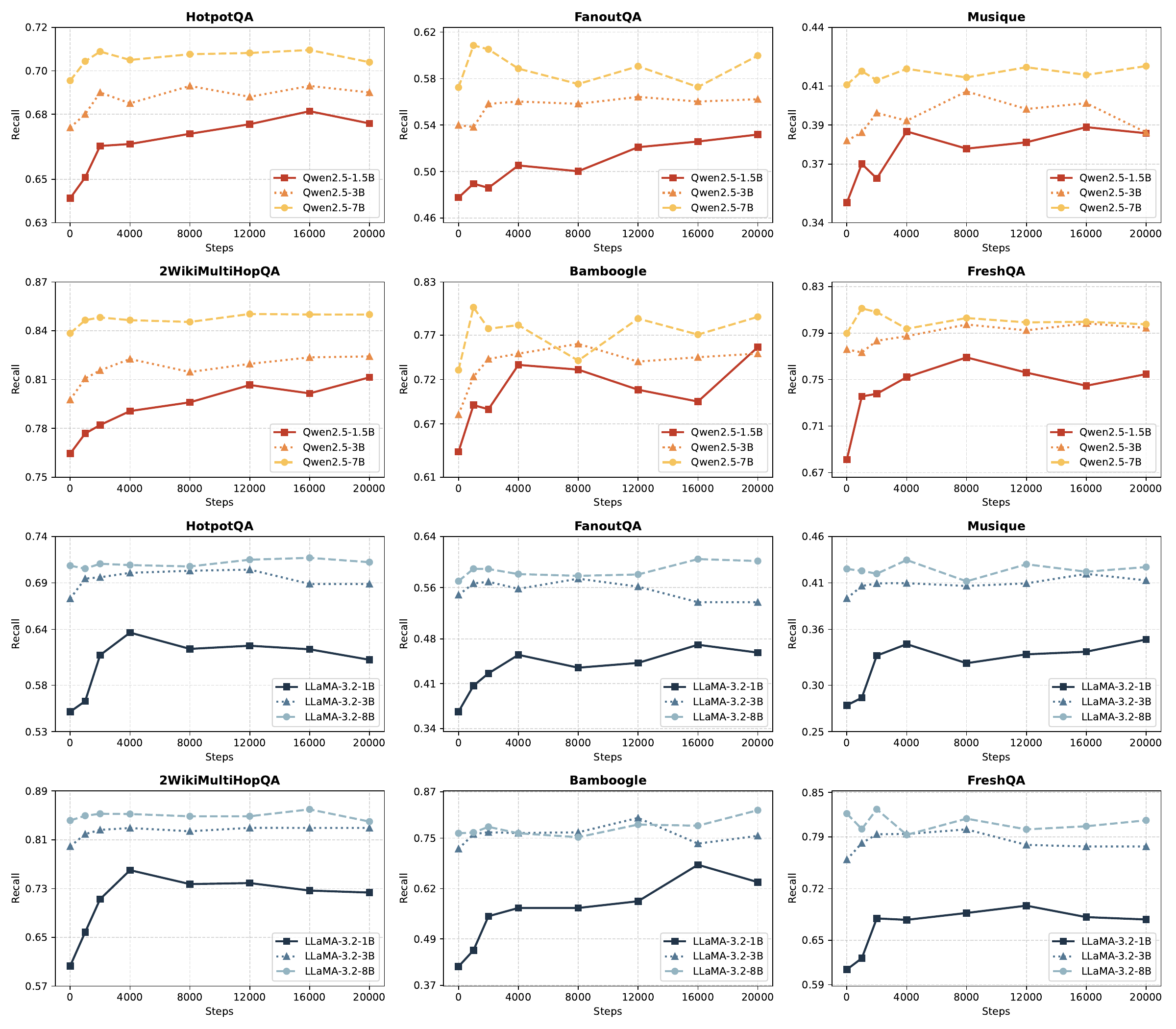}
    \caption{Scaling Performance of SFT with respect to training steps on RAMP. The results obtained at training step 0 align precisely with those of the \textit{Direct SFT} baseline, which is directly fine-tuned on the 58K CoT trajectories derived from HotpotQA.}
    \label{fig:scaling}
\end{figure}

To verify the scaling potential of \masksearch{} as a pre-training task, we conducted experiments using models trained with different numbers of steps in the first stage and evaluated their performance after subsequent finetuning on the downstream task. Our experiments in Figure \ref{fig:scaling} reveal that small models (1B, 1.5B) demonstrate significant improvements in performance after undergoing the pre-training phase. This indicates that the RAMP task is effective in enhancing the agentic search capabilities of models. For larger models, scaling up the dataset size is also effective, but the performance gains are not as pronounced as those observed with smaller models. This can be attributed to the fact that the data used for training the 7B model is generated through a self-evolution process, which may lack diversity and complexity compared to its own prediction. Therefore, the quality and diversity of the training data are critical factors in determining the upper limit of the model's performance during SFT.

\subsection{Supervised Curriculum Learning}

\begin{figure}[t]
    \centering
    \begin{minipage}[t]{0.45\textwidth}
        \centering
        \includegraphics[width=0.8\linewidth]{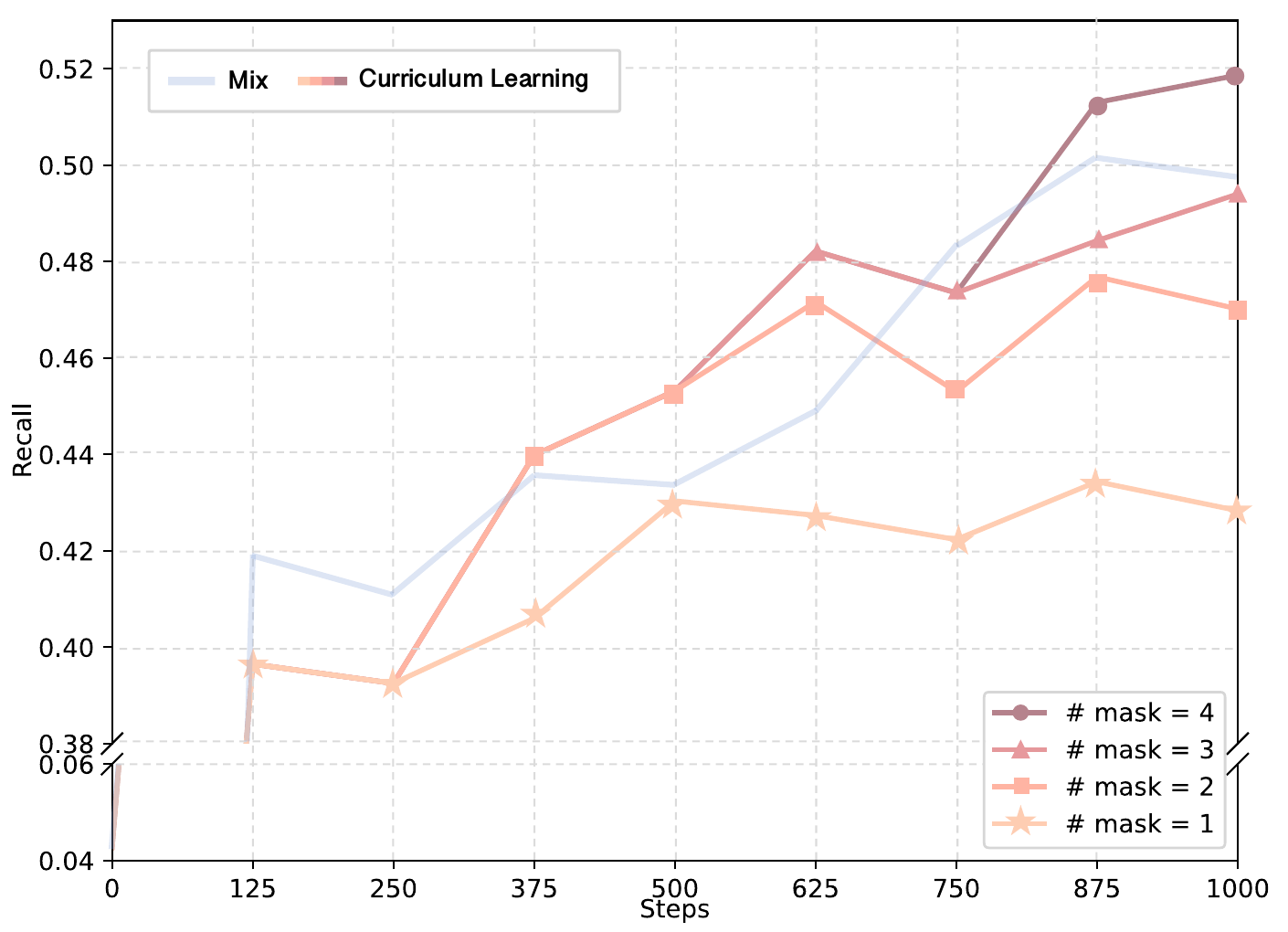}
        \caption{Performance on the dev set while finetuning with varying numbers of masks.}
        \label{fig:cl_sft_mask_num}
    \end{minipage}
    \hfill 
    \begin{minipage}[t]{0.5\textwidth}
        \vspace{-3.5cm} 
        \centering
        \captionof{table}{Average recall of different training strategies on the test datasets.}
        \label{tab:cl_sft_res}
        \resizebox{0.9\textwidth}{!}{
        \begin{tabular}{lccc}
        \toprule
        \multirow{2}{*}{\textbf{Model}} & \multicolumn{2}{c}{\textbf{w/ RAMP}} & \multirow{2}{*}{\textbf{w/o RAMP}}\\
        \cmidrule{2-3}
        & \textbf{CL} & \textbf{Mix} & \\
        \midrule
            \rowcolor{gray!20} 
            \multicolumn{4}{c}{\textit{Qwen}} \\
            Qwen2.5-1.5B & 54.06 & \textbf{55.36} & 52.54\\
            Qwen2.5-7B & \textbf{65.42} & 65.36 & 64.35 \\
            \midrule
            \rowcolor{gray!20} 
            \multicolumn{4}{c}{\textit{LLaMA}} \\
            LLaMA-3.2-1B & \textbf{55.93} & 53.67 & 52.98 \\
            LLaMA-3.1-8B & 64.93 & \textbf{65.57} & 64.21 \\
        \bottomrule
        \end{tabular}}
        
    \end{minipage}
\end{figure}

In this section, we delve deeper into the effectiveness of curriculum learning in the context of RAMP and downstream task training via SFT. We sample 10K data from RAMP for each number of masked spans and 6K from HotpotQA to maintain an appropriate ratio between pre-training and downstream tasks. Additionally, 500 QA pairs are sampled from the remaining RAMP data as a validation set, with 100 data points for each number of masked spans. As illustrated in Figure~\ref{fig:cl_sft_mask_num}, we observe a clear trend where increasing the number of masked spans leads to significant performance improvements on the validation set. Although the initial performance lags behind, the curriculum learning approach ultimately outperforms the mixed training strategy, which is the default training method that mixes all the data together. The advantage observed in the validation set has the potential to carry over to the downstream tasks after fine-tuning. As shown in Table~\ref{tab:cl_sft_res}, CL outperforms the mixed training strategy when using the Qwen2.5-7B and LLaMA-3.2-1B models, indicating that the approach can generalize across different model architectures.

%% file: Mask_QA/section/discussion.tex
\section{Discussion}

In this section, we conduct an in-depth discussion of the critical factors of RAMP and its training process to offer a comprehensive exploration of our approach. A case study is presented in the Appendix \ref{appendix: case} to provide further insights into the practical application of our method.

\subsection{Masking Strategy}

\begin{figure}[t]
  \centering
  \includegraphics[width=0.95\textwidth]{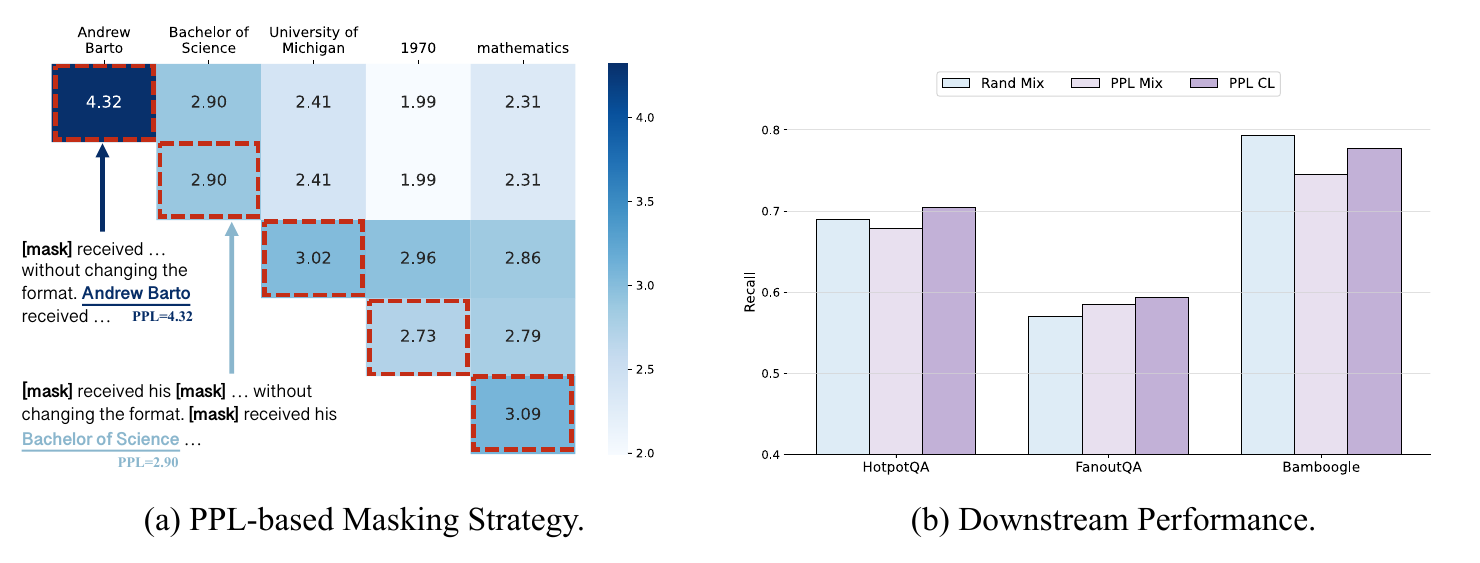}
  \caption{Exploring RAMP with PPL-based masking strategy. The subfigure (a) illustrates the computation of perplexity for the unmasked spans in each round. The subfigure (b) indicates the effect of the PPL-based masking strategy on downstream test sets along with CL.}
  \label{fig:ppl_mask}
\end{figure}

We delve into the impact of masking methods on the RAMP task. Beyond salient span masking, numerous prior studies have investigated the effects of selecting more challenging masks in Masked Language Modeling (MLM), including using Pointwise Mutual Information (PMI) and Perplexity (PPL) to evaluate the difficulty of masked spans. For autoregressive models, we explore the PPL-based masking strategy depicted in Figure \ref{fig:ppl_mask}(a). We measure the difficulty of a masked span to a model by calculating the perplexity, i.e., loss of the span, as it restores the masked span in the original context following the instruction. Specifically, we greedily select the span with the highest PPL among all unmasked salient spans until the required number of masks is met. To validate this strategy against the original random masking strategy, we prepare a 40K RAMP dataset for SFT using our agent-based trajectory construction method and sample 6K HotPotQA trajectories for training in the second stage.

As shown in Figure \ref{fig:ppl_mask}(b), the PPL-based masking strategy yields a performance boost on the FanoutQA dataset, underscoring the efficacy of selecting more challenging masks. However, the experimental outcomes on the HotpotQA and Bamboogle datasets imply that merely augmenting the difficulty of the masked spans may result in a decline in performance. This suggests that the model may not have genuinely mastered the tasks due to its increased difficulty. On the other hand, curriculum learning, which progressively escalates the complexity of the training tasks, proves to be advantageous. After the second stage of training, the model demonstrates enhanced performance across all three datasets, highlighting its effectiveness in enhancing the model's learning progress to tackle intricate tasks.

\subsection{Performance with Different RL Reward}

\begin{figure}[t]
  \centering
  \begin{subfigure}[b]{0.32\textwidth}
    \includegraphics[width=\textwidth]{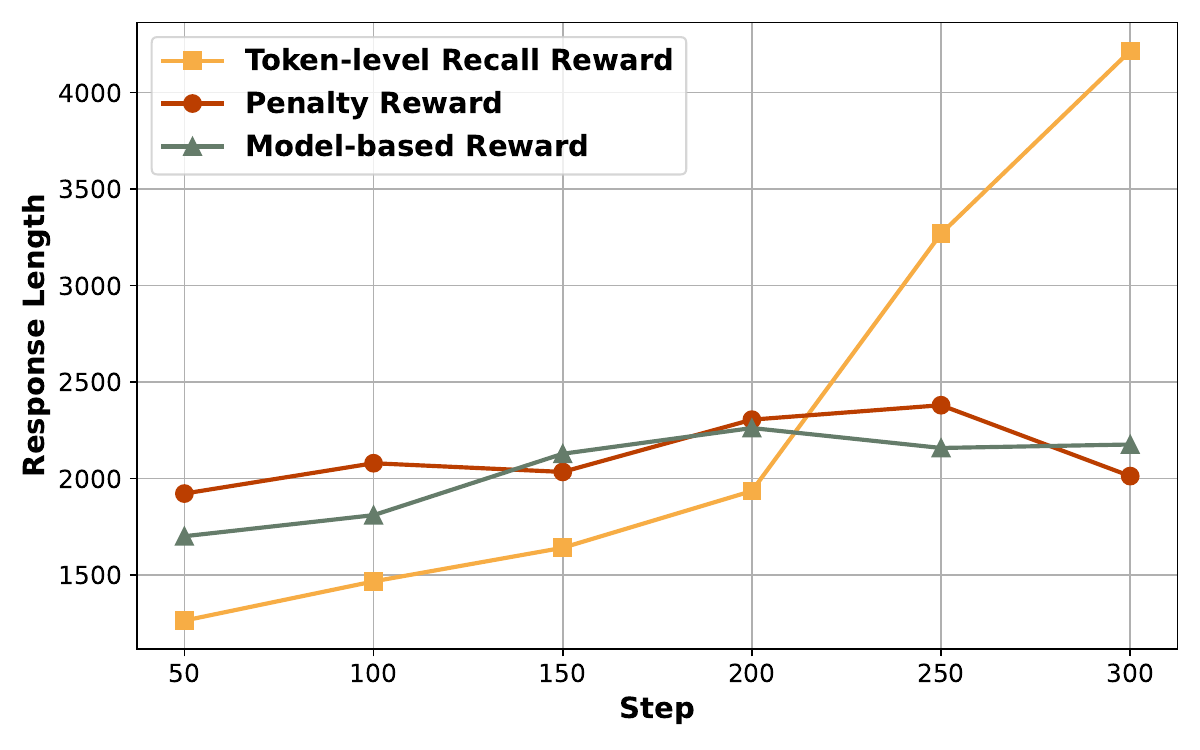}
    \caption{Response Length}
    \label{fig:response_length}
  \end{subfigure}
  \hfill
  \begin{subfigure}[b]{0.32\textwidth}
    \includegraphics[width=\textwidth]{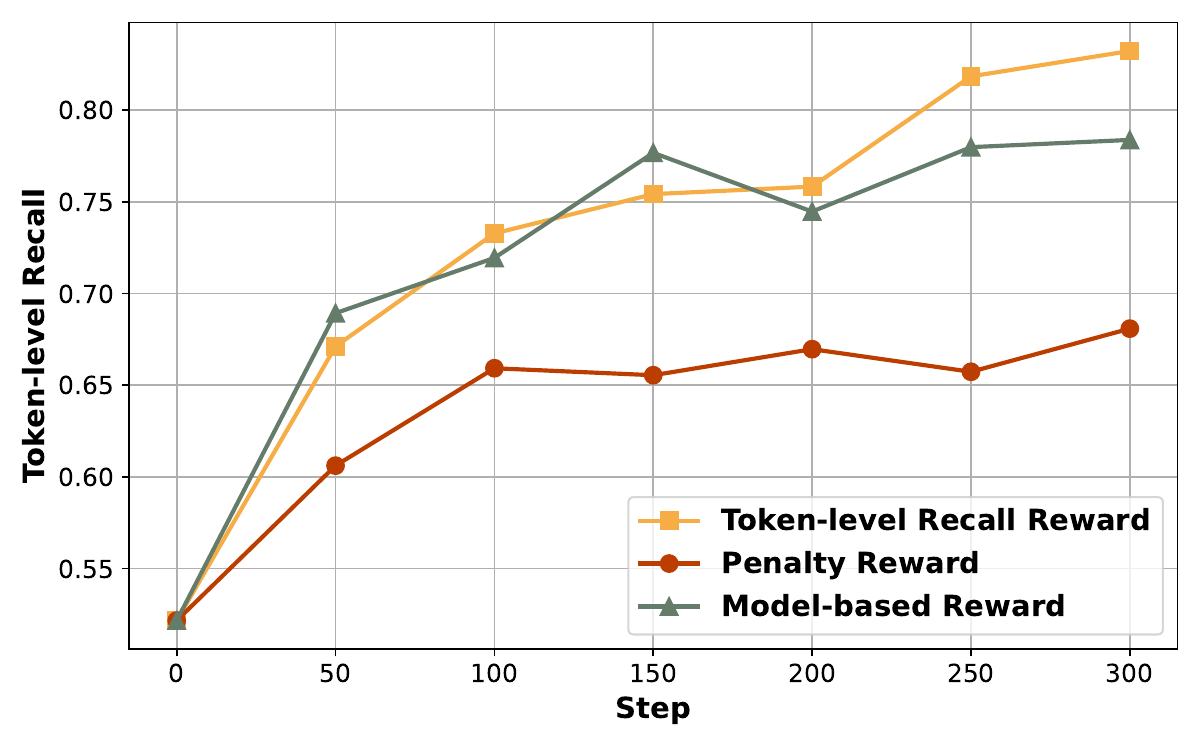}
    \caption{Token-level Recall}
    \label{fig:token_level_recall}
  \end{subfigure}
  \hfill
  \begin{subfigure}[b]{0.32 \textwidth}
    \includegraphics[width=\textwidth]{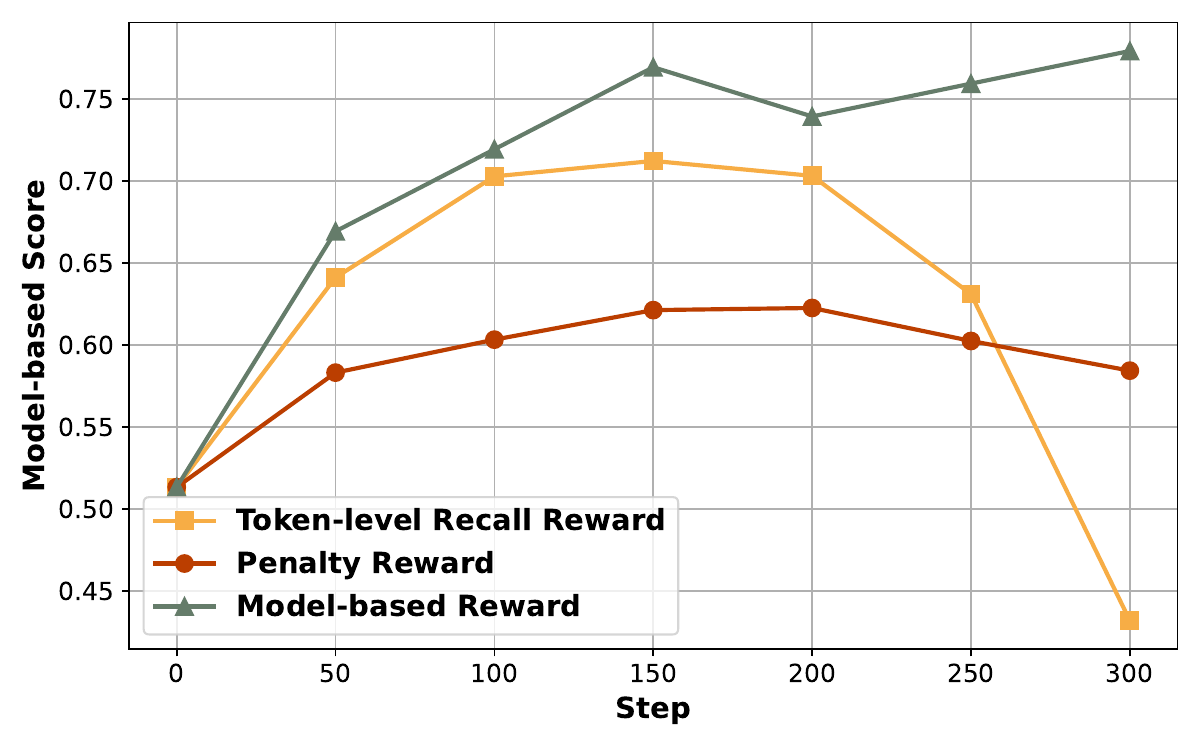}
    \caption{Model-based Score}
    \label{fig:model_based_score}
  \end{subfigure}

  \caption{
  Exploring Different Answer Reward Designs. These figures present the performance of the Qwen-7b model on the development set during RL training. The subfigure (a) illustrates the length of responses generated by the model. The subfigure (b) displays the metric based on token-level recall. The subfigure (c) details the metric assessed using the Qwen-72b model as a judge.
  }
  \label{fig:rl_reward}
\end{figure}

We study the impact of different RL rewards on model performance, as discussed in section \ref{sec:reward}. The results are shown in Figure \ref{fig:rl_reward}, model trained with the token-level recall reward hacks the metric by adding a lot of irrelevant information to the answer, significantly increasing the length of the response. This results in a notable decline in actual performance compared to other RL rewards under model-based evaluation. While penalty-based reward can substantially reduce answer length, performance is still affected and the model still can hack within the constraint of limited answer length by employing enumeration in our observation. Model-based reward offer significant improvements of 34.71 and 19.48 over the other two reward methods, effectively addressing issues of reward hacking and demonstrating remarkable stability and effectiveness in RL training. 
These advantages ultimately led us to adopt the model-based reward approach for training.

%% file: Mask_QA/section/conclusion.tex
\section{Conclusion}

In this paper, we introduce a novel framework \masksearch{} for enhancing the agentic search capabilities of LLMs, specifically through a two-stage training approach. Our method, built around the Retrieval-Augmented Mask Prediction (RAMP) pre-training task, enables models to autonomously perform multi-step search and reasoning to fill in masked spans, fostering a deeper integration of external knowledge. We demonstrate that by using both Supervised Fine-tuning (SFT) and Reinforcement Learning (RL), along with curriculum learning. Our framework leads to substantial performance improvements across both in-domain and out-of-domain open-domain question answering tasks, compared to the baselines. Overall, our work underscores the potential of pre-training in advancing the field of multi-hop reasoning and retrieval-augmented tasks, offering a scalable, transferable framework that enhances the capabilities of LLM-based search agents regardless of model size, architecture, or training methods.

%% file: Mask_QA/appendix/training_setup.tex
\section{Training Settings}

\label{appendix: exp_setup}

\paragraph{Supervised Fine-tuning} During the Continuous Pre-Training (CPT) phase of the scaling experiment, we ensure a global batch size of 1024. For models with 1B, 1.5B, and 3B parameters, we use a distributed training setup with 16 nodes and 8 H20 GPUs per node. For models with 7B and 8B parameters, we use a distributed training setup with 32 nodes and 8 GPUs per node. Each model is trained for 2 epochs with a learning rate set to $4 \times 10^{-5}$. For downstream task training, we use a global batch size of 64 with a learning rate set to $1 \times 10^{-5}$, training on 8 GPUs per node.

\paragraph{Reinforcement Learning} 
For DAPO training, we configure the batch size to 16, set the policy model learning rate to $1 \times 10^{-6}$, and sample 16 responses per prompt. The training for the CPT and downstream tasks is performed on 8 H20 GPUs over 1 epoch and early stops training when there is no better result on the development set for 150 steps. We use the instruct models for RL, as base models often fail to follow the instructions.

%% file: Mask_QA/appendix/ralm.tex
\section{Detailed Comparison with Existing RALMs}

\begin{wraptable}[16]{r}{8.3cm}
\setlength{\tabcolsep}{1pt}
\vspace{-1.3em}
\caption{Comparison with existing RALMs and search-enhanced reasoning models. E2E is short for end-to-end. }
\vspace{-0.2em}
\centering
\small
\resizebox{8.3cm}{!}{\begin{tabular}{l*{4}{c}}
\toprule
 & \textbf{\makecell{\# Retrieval\\Tokens}} & \textbf{\# Models} & \textbf{Retriever}  & \textbf{\makecell{Multi-\\Step}} \\
\midrule
KNN-LM \cite{khandelwal2020generalizationmemorizationnearestneighbor} & $O(10^{9})$ & 2 & Transformer (Frozen)  & \XSolidBrush \\
REALM \cite{guu2020realm}    & $O(10^{9})$ & 2 & BERT (2-Stage E2E)  & \XSolidBrush\\ 
RAG \cite{Lewis2020RetrievalAugmentedGF} & $O(10^{9})$ & 2 &  DPR (E2E)  & \XSolidBrush \\
Retro \cite{borgeaud2022improvinglanguagemodelsretrieving} & $O(10^{12})$ & 2 & BERT (Frozen) & \XSolidBrush \\ 
Atlas \cite{izacard2022atlasfewshotlearningretrieval} & $O(10^{10})$ & 2 & Contriever (E2E) &\XSolidBrush\\
IC-RALM \cite{ram2023incontextretrievalaugmentedlanguagemodels} & $O(10^{8})$ & 2 & BM25 &\XSolidBrush\\ 
RADIT \cite{lin2024raditretrievalaugmenteddualinstruction} & $O(10^{11})$ & 2 & DRAGON+ (Separate) &\XSolidBrush\\ 
\hline
Search-R1 \cite{jin2025searchr1trainingllmsreason} & $\infty$ & 1 & LLM (1-Stage RL) & \Checkmark\\
\masksearch & $\infty$ & 1  & LLM (2-Stage E2E) & \Checkmark \\
\bottomrule
\end{tabular}}
\label{tab:existing}
\end{wraptable}
\setlength{\tabcolsep}{6pt}

\masksearch introduces several innovations that distinguish it from prior work in the field of Retrieval-Augmented Language Models (RALMs) and search-enhanced reasoning models. Firstly, in terms of the number of retrieval tokens, \masksearch operates with an effectively infinite retrieval space, which is a significant departure from traditional models that are limited by the size of their pre-defined retrieval corpora. 
Secondly, \masksearch employs a single-model architecture, in contrast to the dual-model setups of traditional RALMs. This simplification not only reduces computational overhead but also enhances the coherence and consistency of the reasoning process. By interacting with a search engine in agentic style, \masksearch supports multi-step reasoning, a capability that is not present in most of the models listed in the table. This feature is crucial for handling complex queries that require the integration of information from multiple sources and cannot be solved in a single step. 

The work most similar to \masksearch is Search-R1, but there are several key differences that set \masksearch apart. While both models operate with an effectively infinite retrieval space and support multi-step reasoning, \masksearch employs a two-stage end-to-end training process that can be adapted to both SFT and RL. This contrasts with Search-R1, which uses a single-stage reinforcement learning approach for its retriever. The two-stage end-to-end training in \masksearch allows for more general optimization of the agent capabilities, leading to better integration with the LLM and improved performance.

%% file: Mask_QA/appendix/results.tex
\section{Results for LLaMA Models}

In this section, we present the results of our experiments conducted on varying sizes of LLaMA models. Our experiments were specifically carried out on the SFT, as is shown in Table \ref{tab:llama_performance}. The main findings are summarized as follows:

\paragraph{Significant Performance Improvement over Baselines} Similar to Qwen models, our experiments demonstrated that the \masksearch framework achieved substantial performance improvements across multiple datasets compared to existing methods. On the LLaMA-3.2-1B model, \masksearch achieved an average performance improvement of nearly 10 points over the next best method (Distilled Search-R1). This significant enhancement highlights the effectiveness of \masksearch in improving search capabilities.

\paragraph{Scalability Across Model Sizes} The \masksearch framework showed consistent performance improvements across different sizes of LLaMA models, ranging from 1B to 8B parameters. However, the extent of improvement for the SFT stage depends on the capability gaps between the distilled model and the student model, which is also observed with the Qwen series. 

\paragraph{Transferability among Different Model Series} The \masksearch framework showed strong transferability among different architectures of models, as we utilize data generated by the Qwen-series model but still achieve significant performance improvements for LLaMA models. This demonstrates the framework's ability to leverage data from different sources and adapt to different model architectures, further enhancing its applicability in diverse research and development scenarios. 

\begin{table}[htbp]
    \centering
    \caption{Complete results of LLaMA models.}
    \resizebox{1.0\textwidth}{!}{
    \begin{tabular}{lccccccccc}
    \toprule[0.05pt]
    \toprule[0.05pt]
        \textbf{Methods} &  \textbf{Pre-Training} &  \textbf{Post-Training} & \textbf{HotpotQA} & \textbf{FanoutQA} & \textbf{Musique} & \textbf{2Wiki} & \textbf{Bamboogle} & \textbf{FreshQA} & \textbf{Avg.} \\
    \midrule
        \rowcolor{gray!20} 
        \multicolumn{10}{c}{\textit{LLaMA-3.2-1B}} \\
        RAG-PE & \XSolidBrush & \XSolidBrush & 20.00 & 29.91 & 9.03 & 40.15 & 13.23 & 42.74 & 32.51 \\
        Agent-PE & \XSolidBrush & \XSolidBrush & 37.51 & 31.73 & 19.14 & 45.96 & 32.22 &  57.24 & 37.30 \\
        Distilled Search-R1 & \XSolidBrush & SFT & 55.15 & 36.62 & 27.84 & 60.32 & 41.87 & 61.05 & 47.14 \\
         \midrule
        \masksearch & SFT & SFT & 63.66 & 45.50 & 34.41 & 76.01 & 56.99 & 67.77 & 57.40 \\
    \midrule
    \rowcolor{gray!20} 
        \multicolumn{10}{c}{\textit{LLaMA-3.2-3B}} \\
        RAG-PE & 34.00 & 48.84 & 15.44 & 53.79 & 46.35 & 63.33 & 48.63 \\
        Agent-PE & 51.90 & 37.17 & 42.66 & 71.44 & 48.46 & 58.06 & 51.61 \\
        Distilled Search-R1 & \XSolidBrush & SFT & 67.31 & 54.86 & 39.33 & 79.91 & 72.25 & 75.91 & 64.93 \\
         \midrule
        \masksearch & SFT & SFT  & 70.31 & 57.40 & 40.67 & 82.39 & 76.52 & 80.00 & 67.88 \\
    \midrule
    \rowcolor{gray!20} 
        \multicolumn{10}{c}{\textit{LLaMA-3.1-8B}} \\
        RAG-PE & 36.67 & 52.38 & 19.74 & 51.52 & 51.04 & 63.33 & 50.06 \\
        Agent-PE & 59.70 & 52.30 & 41.12 & 77.09 & 59.69 & 71.85 & 60.29 \\
        Distilled Search-R1 & \XSolidBrush & SFT & 70.85 & 57.01 & 42.50 & 84.15 & 76.27 & 82.13 & 68.82 \\
         \midrule
        \masksearch & SFT & SFT  & 71.70 & 60.46 & 42.20 & 85.98 & 78.22 & 80.42 & 69.83 \\
    \bottomrule[0.05pt]
    \bottomrule[0.05pt]
    \end{tabular}
    }
    \label{tab:llama_performance}
\end{table}

%% file: Mask_QA/appendix/prompt.tex
\label{sec: ma_prompt}

This section presents the prompts used for agent-based 
reasoning trajectories construction, including the planner agent, rewriter agent, observer agent and LLM-Judge.

\setlength{\abovecaptionskip}{0pt}
\begin{longtable}{>{\raggedright\arraybackslash}p{0.96\linewidth}}

\caption{Prompt for Agent-Based CoT trajectory construction.}
\label{tab:agent_prompt} \\
\toprule
\endhead

\midrule
\multicolumn{1}{r}{\textit{Continued on next page}} \\
\endfoot
\bottomrule
\endlastfoot

\textbf{Planner Agent}\\
\midrule
Your task is to provide the steps for solving a multi-hop search problem. 

The output format should be: 
"\think{\{Overall thought process\}}\\\search{[\{"query": "{query}", "intent": 1\}]}"; 
query should be in sentence format.

\\
Here are some examples:

\\
\textbf{Question:} What is the undergraduate school of the director of the movie "Sense and Sensibility"?

\textbf{Thought:} \think{To answer this question, I will take the following steps:\\1. First, find out who directed the movie "Sense and Sensibility".\\2. Investigate the educational background of the director, particularly their undergraduate school.\\3. Identify the specific institution where the director completed their undergraduate studies.\\Now, I will start with the first step and search for the director of the movie "Sense and Sensibility".}\\\search{[\{"query": "Who is the director of the movie 'Sense and Sensibility'?", "intent": 1\}]}

\\

\textbf{Question:} When did the birthplace of the performer of Live and Beyond become the capital of the state where Knowles is located?

\textbf{Thought:} \think{To determine when the birthplace of the performer of Live and Beyond became the capital of the state where Knowles is located, I will take the following steps: \\1. First, identify who the performer of Live and Beyond is.\\2. Then find out the birthplace of this performer.\\3. Next, search for which state Knowles is from.\\4. Finally, determine when the birth city of the performer of Live and Beyond became the capital of Knowles' state.\\Now, I will start with the first step and search online to determine who the performer of Live and Beyond is.}\\\search{[\{"query": "Find out who the performer of Live and Beyond is", "intent": 1\}]}

\\

\textbf{Question:} \{input\}\\
\textbf{Thought:}

 \\
 
\toprule
\textbf{Rewriter Agent} \\
\midrule

Given a piece of content containing queries to search, your task is to rewrite the queries in order to obtain more comprehensive search results. Please provide at least three rewritten queries.

The output format should be the following JSON structure: \\

\{"queries": ["query 1", "query 2", "query 3"]\}

\\
Here are some examples:

\\

\textbf{Content:}
\think{To find out which undergraduate school the director of the movie "Sense and Sensibility" attended, I will take the following steps: \\
1. First, determine who the director of the movie "Sense and Sensibility" is. \\
2. Then, search for educational background of this director, particularly undergraduate education. \\

Now, I'll proceed with the first step by using online searches to identify the director of the movie "Sense and Sensibility".}\\
\search{[\{"query": "Find out who the director of the movie 'Sense and Sensibility' is", "intent": 1\}]}

\textbf{Rewritten Queries:} \\

\{"queries": ["Sense and Sensibility director", "Sense and Sensibility 1995 director", "Sense and Sensibility Filmmaker"]\}

\\

\textbf{Content:}
\think{After analyzing the search results in detail, I concluded that the director of the movie "Sense and Sensibility" is Ang Lee. Therefore, I will proceed with the next step, where I need to search for his detailed undergraduate education.} \\
\search{
[\{"query": "Search for the undergraduate school of Ang Lee", "intent": 1\}]}

\textbf{Rewritten Queries:} \\

\{"queries": ["Ang Lee education background", "Ang Lee undergraduate school", "Ang Lee biography"]\}

\\
\textbf{Content:} \{input\}\\
\textbf{Rewritten Queries:} \\

\\
\toprule
\textbf{Observer Agent} \\
\midrule

Given a multi-hop search problem, the steps already taken, and the search results obtained from the last executed step, your task is to:
\\

1. Carefully analyze the search results to determine if they resolve the previous step, and provide a summary.\\
2. Update the execution steps and propose new steps based on the search results. You may encounter the following scenarios:\\
   \hspace{0.5cm} a. If the current search results are sufficient to arrive at the Final Answer to the problem, provide the final answer directly. Use the format: "\think{\{thought process\}}\answer{\{answer\}}". Please strictly adhere to this format.\\
   \hspace{0.5cm} b. If the current search results resolve the previous step, proceed according to the original plan and provide the next new step. Use the format: "\think{\{thought process\}}\\\search{[\{"query": "query", "intent": 1\}]}"; the query should be in sentence format.\\
   \hspace{0.5cm} c. If the current search results do not resolve the previous step, reflect on and update the previous execution steps based on the search results. Use the format: "\think{\{reflection\}}\search{[\{"query": "query", "intent": 1\}]}"; the query should be in sentence format.\\
   \hspace{0.5cm} d. If after multiple reflections (at least twice), the search still fails to resolve the issue, supplement based on your knowledge and provide the next new step. Use the format: "\think{\{thought process\}}\search{[\{"query": "query", "intent": 1\}]}"; the query should be in sentence format.

\\
Below are some examples:

\\
\textbf{Question:} What is the undergraduate school of the director of the movie "Sense and Sensibility"?\\
\textbf{Thought:} \think{To find out the undergraduate school of the director of the movie "Sense and Sensibility", I will take the following steps:\\1. First, determine who directed the movie "Sense and Sensibility". \\2. Then, search for this director’s educational background, focusing on undergraduate education. \\I will now begin with step one by searching online to identify the director of the movie "Sense and Sensibility".}\\
\search{[\{"query": "Find out who directed the movie 'Sense and Sensibility'", "intent": 1\}]}\\
\textbf{Observation}: \information{
\\1. The movie "Sense and Sensibility" was released in 1995 and is directed by Ang Lee, a renowned director known for his work in both Eastern and Western cinema.\\
2. Ang Lee is recognized for directing major films such as "Crouching Tiger, Hidden Dragon" and "Life of Pi".\\
3. Ang Lee's background in filmmaking and storytelling has earned him multiple Academy Awards and international acclaim.}\\
\textbf{Thought}: \think{After a detailed analysis of the search results, I concluded that Ang Lee is the director of the movie "Sense and Sensibility". Thus, I will proceed to the next step and search for Ang Lee’s educational background, focusing on his undergraduate education.}\\
\search{[\{"query": "Search for the undergraduate school of Ang Lee", "intent": 1\}]}

\\

\textbf{Question:} What is the undergraduate school of the director of the movie "Sense and Sensibility"?\\
\textbf{Thought:} \think{To find out the undergraduate school of the director of the movie "Sense and Sensibility", I will take the following steps:\\1. First, determine who directed the movie "Sense and Sensibility". \\2. Then, search for this director’s educational background, focusing on undergraduate education. \\I will now begin with step one by searching online to identify the director of the movie "Sense and Sensibility".}\\
\search{[\{"query": "Find out who directed the movie "Sense and Sensibility", "intent": 1\}]}\\
\textbf{Thought}: \think{After a detailed analysis of the search results, I concluded that Ang Lee is the director of the movie "Sense and Sensibility". Thus, I will proceed to the next step and search for Ang Lee’s educational background, focusing on his undergraduate education.}\\
\search{[\{"query": "Search for the undergraduate school of Ang Lee", "intent": 1\}]}
\textbf{Observation}: \information{\\
1. Ang Lee graduated from the National Taiwan College of Arts in 1975 and then came to the U.S. to receive a B.F.A. Degree in Theatre/Theater Direction at the University of Illinois at Urbana-Champaign, and a Masters Degree in Film Production at New York University.\\
2. After high school Lee attended the Taiwan Academy of Art, where he became interested in acting. He moved to the United States in 1978 to study theater at the University of Illinois at Urbana-Champaign.}\\
\textbf{Thought}: \think{After a thorough analysis of the search results, I found out that Ang Lee graduated from the National Taiwan University with a degree in French literature, as well as a B.F.A. Degree from the University of Illinois at Urbana-Champaign, therefore, the answer is} \answer{National Taiwan College of Arts and University of Illinois at Urbana-Champaign.}

\\

\textbf{Question}: Who is the director of the suspense drama that premiered on MBC?\\
\textbf{Thought}: \think{To solve this problem, I need to follow a few steps in my search: \\1. First, I need to identify which suspense drama premiered on MBC.\\2. Then, I will look up who directed that suspense drama.\\First, I will search for recent suspense dramas that premiered on MBC.}\\\search{[\{ "query": "What recent suspense dramas premiered on MBC?", "intent": 1\}]}
\\
\textbf{Observation}: \information{
1. Premiering soon! A 12-episode suspense drama is coming, starring Wang Zhiwen and Yu Ai... From the trailer, 'The Vanishing Elephant' combines suspense and comedy, bringing an amusing story with humor and absurdity.

2. Chen Weiting's suspense drama is airing, featuring time travel and criminal investigation, with top actors like award-winning actors joining."

3. The best 10 suspense dramas of the past 4 years, with 'The Big Squad' at the bottom. Though the series is framed as a historical drama, it unfolds as a large-scale suspense espionage drama.

4. Premiering soon! An 18-episode suspense drama is coming, starring Sun Li, with a strong cast. However, recent suspense dramas such as 'Dim Flames' and 'Rebirth' have not been well-received.

5. Lee Jong-suk's return! New drama 'The Black Lawyer' premiered on MBC, achieving high viewership. The story is a suspenseful crime drama about a vulnerable lawyer who is framed and battles powerful forces.

6. The top 10 suspense dramas to watch in 2024, featuring leading actors like Liao Fan, Duan Yihong, and Zhang Songwen...

7. Premiering tonight! Another suspenseful drama is coming! 'Thirteen Years in Dust' will air on iQIYI's Mist Theater, starring Chen Jianbin and Chen Xiao.

8. Chen Weiting's new drama 'The Sun and Stars' debuted with impressive suspenseful plotlines.
}\\
\textbf{Thought}: \think{From the search results, I found several relevant pieces of information:\\1. The suspense drama 'The Black Lawyer' premiered on MBC, starring Lee Jong-suk and Im Yoon-ah, and it is a crime suspense drama. \\2. The premiere of 'The Black Lawyer' achieved high ratings, telling a story about a vulnerable lawyer who is framed and fights back against powerful figures. \\3. However, the search results did not provide the director's information for 'The Black Lawyer.' \\Filter out unrelated content: The search results also mention suspense dramas on other platforms like iQIYI and Mist Theater, and other non-MBC dramas, such as 'The Sun and Stars.' These are not directly related to the question and can be excluded. \\Current progress summary: Based on the retrieved information, 'The Black Lawyer' is a recent suspense drama premiered on MBC, but the director’s information is still missing. \\Next plan: I will continue with the second step of my plan, searching for the director of 'The Black Lawyer.'}\\\search{[\{"query": "Who is the director of 'The Black Lawyer'?", "intent": 1\}]}

\textbf{Question:} \{input\}\\
\textbf{Thought:} \{thought\}\\
\textbf{Observation:} \{observation\}\\
\textbf{Thought:}

\\
\toprule
\textbf{LLM-Judge}\\
\midrule
Given the correct answer to a question, determine if the model\'s response is correct. If correct, output "correct"; if incorrect, output "incorrect". Do not include unrelated content.\\
\\
\textbf{Question:} \{question\}

\textbf{Correct Answer:} \{answer\}\\\textbf{Model Response:} \{model\_response\}
\end{longtable}

%% file: Mask_QA/appendix/RL_appendix.tex
\subsection{Training Template}
\begin{table*}[h]
\caption{Training template. The question is appended at the end during RL training and inference.}
\centering
\begin{tabular}{p{12cm}}  
\toprule
Answer the given question. \
You must conduct reasoning inside \think{and} first every time you get new information. \
After reasoning, if you find you lack some knowledge, you can call a search engine by \search{query}, and it will return the top searched results between \information{and}. \
You can search as many times as you want. \
If you find no further external knowledge needed, you can directly provide the answer inside \answer{and} without detailed illustrations. For example, \answer{xxx}. Question:\\
\bottomrule
\end{tabular}
\label{tab:rl_template} 
\end{table*}
As illustrated in Table \ref{tab:rl_template}, during the RL process, we follow \cite{jin2025searchr1trainingllmsreason} by utilizing a multi-round interactive template to guide the reasoning of the policy model. Specifically, the model engages in internal reasoning within the <think> tag, where it analyzes the problem and assesses the information collected. If additional evidence is required, the search query is refined within the <search> tag. Once sufficient information is gathered, the answer is provided within the <answer> tag.

\subsection{RL Algorithms Detail}

\paragraph{DAPO with Search Engine}
Decouple Clip and Dynamic sAmpling Policy Optimization (DAPO) \cite{yu2025dapoopensourcellmreinforcement} is an advanced RL algorithm that enhances the Group Relative Policy Optimization (GRPO) \cite{DBLP:deepseekmath} by incorporating techniques such as dynamic sampling and token-level policy gradient loss. DAPO samples a group of outputs $\{o_i\}_{i=1}^G$ for each question $q$ paired with the answer $a$, and optimizes the policy model via the following objective function:

\begin{equation}
\begin{aligned}
\mathcal{J}_{\text{DAPO}}(\theta) =\quad& \mathbb{E}_{(q,a)\sim \mathcal{D}, \{o_i\}_{i=1}^G\sim \pi_{\theta_\text{old}}(\cdot\mid q)}\\&
\Bigg[\frac{1}{\sum_{i=1}^{G}|o_i|}\sum_{i=1}^{G}\sum_{t=1}^{|o_i|} 
\min \Big( r_{i,t}(\theta) \hat{A}_{i,t},  
\ \text{clip} \Big( r_{i,t}(\theta), 1 - {\varepsilon_{\text{low}}}, 1 + {\varepsilon_{\text{high}}} \Big) \hat{A}_{i,t} \Big) \Bigg]
\label{eq:dapoloss}
\end{aligned}
\end{equation}
where
\begin{equation}
    r_{i,t}(\theta)=\frac{\pi_{\theta}(o_{i,t} \mid q, o_{i,<t};\mathcal{R})}{\pi_{\theta_{\text{old}}}(o_{i,t} \mid q,o_{i,<t});\mathcal{R})},\quad\hat{A}_{i,t} = \frac{R_i - \text{mean}(\{R_i\}_{i=1}^G)}{\text{std}(\{R_i\}_{i=1}^G)}.
\label{eq:advantage_calculation}
\end{equation}
Here, $\mathcal{R}$ is the search engine, $\hat{A}_{i,t}$ represent the advantage, ${\varepsilon_{\text{low}}}$ and ${\varepsilon_{\text{high}}}$ are hyperparameters where we set 0.2 and 0.28 respectively. Additionally, DAPO removes KL Divergence to stabilize the generation of long-cot. We mask the retrieved tokens from the search engine, ensuring the policy gradient objective is computed only over LLM-generated tokens.

\subsection{Model-based Reward Prompt}

\begin{longtable}{>{\raggedright\arraybackslash}p{0.96\linewidth}}

\caption{Prompt for model-based reward design.}\\
\toprule
\endhead

\midrule
\multicolumn{1}{r}{\textit{Continued on next page}} \\
\endfoot
\bottomrule
\endlastfoot

\textbf{Reward Model}\\
\midrule

Please evaluate whether the model's response is correct based on the given question, standard answer, and the model's predicted answer. Your task is to rate the result as: \textbf{Correct} or \textbf{Incorrect}.

\subsection*{Correct Responses}
Here are examples of \textbf{Correct} responses:\\
Question: What are Barack Obama's children's names?\\
Standard Answer: Malia Obama and Sasha Obama\\
Model Prediction 1: Malia Obama and Sasha Obama\\
Model Prediction 2: Malia and Sasha\\
Model Prediction 3: Most people would say Malia and Sasha, but I'm not sure and need to confirm.\\
Model Prediction 4: Barack Obama has two daughters, Malia Ann and Natasha Marian, but they are commonly known as Malia Obama and Sasha Obama.
These responses are \textbf{Correct} because:
They fully include the important information from the standard answer.\\
They do not contain any information that contradicts the standard answer.\\
Only the semantic content is considered; language (English or Chinese), case, punctuation, grammar, and order are not important.\\
The presence of vague statements or guesses is acceptable, as long as the standard answer is included and there is no incorrect or contradictory information.

\subsection*{Incorrect Responses}
Here are examples of \textbf{Incorrect} responses:
Question: What are Barack Obama's children's names?\\
Standard Answer: Malia Obama and Sasha Obama\\
Model Prediction 1: Malia\\
Model Prediction 2: Malia, Sasha, Susan, and Sasha Obama or Malia Obama, or Natasha Marian, or Einstein\\
Model Prediction 3: Although I don't know their exact names, I can say that Barack Obama has two children.\\
Model Prediction 4: You might be thinking of Bessie and Olivia. But you should check the latest references for detailed information. Is that the correct answer?\\
Model Prediction 5: Barack Obama's children
These responses are \textbf{Incorrect} because:
They contain factual statements that contradict the standard answer.\\
The answer is empty, restates the question.\\
The answer lists multiple answers, restates the answer.

\subsection*{Special Notes}
Please note the following: \\
The standard answer may contain multiple aspects of the question's response, and within the same aspect, there may be multiple different descriptions, all of which are correct and are given within the same parentheses, connected by commas. For example, consider the question ''What is the name of the social media platforms purchased by Elon Musk?'':\\
Predicted answers ''Twitter,'' ''Twitter, X,'' and ''X'' are all \textbf{Correct}.\\
For standard answers that contain responses to multiple aspects of the question, the model must provide answers to all aspects to be considered correct; otherwise, it is directly judged as \textbf{Incorrect}. There is no such output as \textbf{Partially Correct}. These answers will be given in different parentheses. For example, consider the question ''Who are the original members of the band The Beatles?'':\\
Predicted answers ''John Lennon, Paul McCartney, George Harrison, Ringo Starr'' that include all answers are considered \textbf{Correct}.\\
Predicted answers like ''John Lennon, Paul McCartney'' that do not include all answers are considered \textbf{Incorrect}.

\subsection*{Additional Guidelines}
Also, pay special attention to the following: \\
For questions with numerical standard answers, the predicted answer should match the standard answer. For example, consider the question ''What is the total length of the Jinshan Railway Huangpujiang Special Bridge in meters?'':\\
Predicted answers ''3518,'' ''3518.1,'' and ''3518.17'' are all \textbf{Correct}.\\
Predicted answers ''3520'' and ''3600'' are all \textbf{Incorrect}.\\
If the model's prediction does not directly answer the question and attempts to bypass or fails to directly provide the standard answer, it is considered an \textbf{Incorrect} answer.\\
If the standard answer contains more information than the question, the predicted answer only needs to include the information mentioned in the question.\\
If it is obvious from the question that the predicted answer has omitted information, it is considered \textbf{Correct}.\\
If it is clear that different translation versions of a name refer to the same person, they are also considered \textbf{Correct}.\\
You should focus more on the match between the standard answer and the model's prediction, rather than whether the standard answer is correct.

\subsection*{Example Question}
Here is a new example question. Please rate the predicted answer as one of the following: \\
Question: \{question\} \\
Standard Answer: \{target\} \\
Predicted Answer: \{predicted answer\}  \\
Only return the option represented by \textbf{Correct} or \textbf{Incorrect}, that is, only return A or B, without adding any other text.

\end{longtable}

%% file: Mask_QA/appendix/case.tex
\section{RAMP Case Study}

\label{appendix: case}

We illustrate how the model acquires the ability to retrieve and reason using the Retrieval-Augmented Mask Prediction Task (RAMP) through a case study presented in Table \ref{tab:ramp_case}. In this example, information about the league, team, and university that David Henry Hoelscher is associated with is masked. The model autonomously utilizes the search engine to reason and fill in the missing information.

Initially, the model establishes a search plan, which includes details about the individual's career, team, and university. In the first step, it retrieves information related to his career and team. After acquiring the initial search results, the model identifies the league associated with the individual based on the retrieved data and initiates another search to confirm the team he played for. Following the second search, the model successfully infers his team affiliation and discovers that the previous search results include his college information. It then requests another search to verify this data. Upon completing the third search, the search agent effectively gathers all required information and fills in the masks.

Through the RAMP task, the model not only learns to perform retrievals but also masters the use of retrieved information for further reasoning and verification until the complete answer is obtained.

\setlength{\abovecaptionskip}{0pt}
\begin{longtable}{>{\raggedright\arraybackslash}p{0.96\linewidth}}

\caption{RAMP Case.}
\label{tab:ramp_case} \\
\toprule
\endhead

\midrule
\multicolumn{1}{r}{\textit{Continued on next page}} \\
\endfoot
\bottomrule
\endlastfoot

\textbf{Question}\\
\midrule
David Hoelscher David Henry Hoelscher (born November 27, 1975) is a former American football defensive tackle. He played one game in the [mask] for the [mask] in 1998. He played college football at [mask].\\Fill in all the [mask] and output the whole paragraph without changing its format.\\

\midrule
\textbf{Response} \\
\midrule

\think{To fill in the blanks in the provided sentence about David Hoelscher, I will perform the following searches:\\1. First, search for David Hoelscher's brief NFL career to determine the specific league and team he played for in 1998.\\2. Then confirm the name of the college where he played football before entering the NFL.\\Now, I will start with the first step and search for David Hoelscher's NFL career details. }\\

\search{[\{"query": "David Hoelscher one game NFL" "David Henry Hoelscher NFL team 1998" "David Hoelscher defensive tackle one game in 1998", "intent": 1\}]\\} 

\information{
\\1. David Hoelscher, , - Career Stats - NFL - CBSSports.com: Get the latest career stats for David Hoelscher of the on CBS Sports.\\
2.  David Hoelscher, , - 2021-22 Game Log - NFL - CBSSports.com: Get up-to-date stats for every game played by David Hoelscher during the 2021-22 NFL season on CBS Sports.\\
3. David Hoelscher Stats, News and Video - DE | NFL.com: David Hoelscher DE Player Info Height 6-6 Weight 261 Arms Hands Experience 2 College Eastern Kentucky. \\
4. David Hoelscher - 2021-22 - Esports - Florida Southern College: David Hoelscher. Class Freshman. Hometown Orchard Park, N.Y. \\
5. David Hoelscher | American Football Database - Fandom: American football defensive tackle in the National Football League for the Washington Redskins. He played college football at Eastern Kentucky University. \\
}\\

\think{After finding out that David Hoelscher is indeed associated with the National Football League (NFL) and a specific team, I will now confirm the name of the team he played for in 1998 when he played one game.\\}\\

\search{[\{"query": "David Hoelscher one game in 1998 NFL team" "Which team did David Hoelscher play for in one game in 1998" "David Hoelscher NFL team when he played college football at Eastern Kentucky", "intent": 1\}]\\}

\information{
\\1. David Hoelscher 1998 Game Log | StatMuse: David Hoelscher's game log for the 1998 NFL regular season and playoffs. TEAM, W, L, T, PCT. Eagles NFL 2024 Betting. Best Record ATS · Chargers. 12-5-1.\\
2. David Hoelscher Stats, News and Video - DE | NFL.com: David Hoelscher DE Player Info Height 6-6 Weight 261 Arms Hands Experience 2 College Eastern Kentucky.\\
3. David Hoelscher - Washington Redskins Defensive Lineman: David Hoelscher played 1 season for the Redskins. He played but didn't record a tackle. \\
4. Colonels in the NFL - Eastern Kentucky University Athletics: Colonels in the NFL; David Hoelscher (1994-97), Washington Redskins (1998); Joe Hollingsworth (1946-47), Pittsburgh Steelers (1949-51); Tyrone\\
5. David Hoelscher Pro Football Stats, Position, College, Draft: DAVID HOELSCHER; 1998 Washington Redskins (NFL), 91, DT, 1, 0; 1999 Frankfurt Galaxy (NFLE), 91, RDT-RDE.\\
}\\

\think{After confirming that David Hoelscher played for the Washington Redskins in 1998, I will now verify if there is any additional confirmation needed about the college he attended before entering the NFL, which was mentioned as Eastern Kentucky in the previous searches.\\}\\

\search{[\{"query": "David Hoelscher college football" "Which college did David Hoelscher play football before entering the NFL?" "David Hoelscher defensive tackle college", "intent": 1\}]\\}

\information{
\\1. David Hoelscher Pro Football Stats, Position, College, Draft: DAVID HOELSCHER. Hoelscher, David Henry Height: 6-6 Weight: 256. Born: November 27, 1975 Coldwater, OH High School: Versailles (OH). Year, College, Status. 1994.\\
2. HOELSCHER; 1994, Eastern Kentucky, Roster; 1995, Eastern Kentucky, Roster; 1996, Eastern Kentucky, Roster; 1997, Eastern Kentucky, Played. \\
3. David Hoelscher - 2021-22 - Esports - Florida Southern College: David Hoelscher. Class Freshman. Hometown Orchard Park, N.Y.. Prev School Athletics. 111 Lake Hollingsworth Drive | Lakeland FL 33801 | 863-680-4245. \\
4.David Hoelscher Stats, Height, Weight, Position, Draft, College: Position: DE; Born: November 27, 1975 in Coldwater, OH; College: Eastern Kentucky; Weighted Career AV (100-95-): 0 (17977th overall since 1960); High School:.\\
5. David Hoelscher | American Football Database - Fandom: He played college football at Eastern Kentucky University. Stub icon, This biographical article relating to an American football defensive lineman born in the.
\\
}\\

\think{After verifying the college where David Hoelscher played football, I can now complete the original statement. \\}\\

\answer{
David Hoelscher David Henry Hoelscher (born November 27, 1975) is a former American football defensive tackle. He played one game in the National Football League for the Washington Redskins in 1998. He played college football at Eastern Kentucky University.
}
\end{longtable}

%% file: Mask_QA/appendix/Impacts_Ethics.tex
\section{Broader Impacts} \label{appendix:impact}
As an exploration in the field of Search Agents, \masksearch{} has the following impacts:
\begin{itemize}
    \item Provides a general pre-training method for Search Agent: \masksearch{} introduces the Retrieval Augmented Mask Prediction Pre-training Task, pioneering the enhancement of an agent's retrieval and reasoning capabilities during the pre-training stage. This offers new perspectives for future developments in agent technology.
    \item Applications in Various RAG Fields: As a general pre-training task, it can be broadly applied to RAG applications across different knowledge domains.
    \item  Negative social impacts: There are no negative social impacts foreseen.
\end{itemize}

\section{Limitations} \label{appendix:limitation}
Despite the empirical success and intuitive motivation of our approach, there are several limitations that warrant further investigation. First, we employ only a search tool for knowledge retrieval to adhere to the concept of RALM. However, agents are capable of utilizing a diverse array of tools, and we believe that the RAMP task could be generalized to incorporate the use of multiple tools. Future work could explore the diversity of toolset, potentially expanding the application scope beyond open-domain QA to other scenarios. Furthermore, while our method has demonstrated promising results, a more in-depth theoretical analysis is necessary to fully understand the factors contributing to its effectiveness.

\section{Data Ethics Statement}
In this paper, we conduct experiments using publicly available datasets, including Hotpot, FanoutQA, Musique, 2WikiMultiHopQA, Bamboogle, and FreshQA, in accordance with their respective usage terms and conditions.

%% file: neurips_2025.bbl
\begin{thebibliography}{61}
\providecommand{\natexlab}[1]{#1}
\providecommand{\url}[1]{\texttt{#1}}
\expandafter\ifx\csname urlstyle\endcsname\relax
  \providecommand{\doi}[1]{doi: #1}\else
  \providecommand{\doi}{doi: \begingroup \urlstyle{rm}\Url}\fi

\bibitem[Yang et~al.(2025)Yang, Yang, Zhang, Hui, Zheng, Yu, Li, Liu, Huang, Wei, et~al.]{qwen2025qwen25technicalreport}
An~Yang, Baosong Yang, Beichen Zhang, Binyuan Hui, Bo~Zheng, Bowen Yu, Chengyuan Li, Dayiheng Liu, Fei Huang, Haoran Wei, et~al.
\newblock Qwen2.5 technical report.
\newblock \emph{arXiv preprint arXiv:2501.15383}, 2025.
\newblock URL \url{https://arxiv.org/abs/2412.15115}.

\bibitem[DeepSeek-AI et~al.(2025)DeepSeek-AI, Guo, Yang, Zhang, Song, Zhang, Xu, Zhu, et~al.]{DeepSeekAI2025DeepSeekR1IR}
DeepSeek-AI, Daya Guo, Dejian Yang, Haowei Zhang, Jun-Mei Song, Ruoyu Zhang, Runxin Xu, Qihao Zhu, et~al.
\newblock Deepseek-r1: Incentivizing reasoning capability in llms via reinforcement learning.
\newblock \emph{ArXiv}, abs/2501.12948, 2025.
\newblock URL \url{https://api.semanticscholar.org/CorpusID:275789950}.

\bibitem[Dubey et~al.(2024)Dubey, Jauhri, Pandey, Kadian, Al-Dahle, Letman, Mathur, Schelten, Yang, Fan, et~al.]{grattafiori2024llama3herdmodels}
Abhimanyu Dubey, Abhinav Jauhri, Abhinav Pandey, Abhishek Kadian, Ahmad Al-Dahle, Aiesha Letman, Akhil Mathur, Alan Schelten, Amy Yang, Angela Fan, et~al.
\newblock The llama 3 herd of models, 2024.
\newblock URL \url{https://arxiv.org/abs/2407.21783}.

\bibitem[Minaee et~al.(2024)Minaee, Mikolov, Nikzad, Chenaghlu, Socher, Amatriain, and Gao]{Minaee2024LargeLM}
Shervin Minaee, Tom{\'{a}}s Mikolov, Narjes Nikzad, Meysam Chenaghlu, Richard Socher, Xavier Amatriain, and Jianfeng Gao.
\newblock Large language models: A survey.
\newblock \emph{ArXiv}, abs/2402.06196, 2024.
\newblock URL \url{https://api.semanticscholar.org/CorpusID:267617032}.

\bibitem[Chang et~al.(2024)Chang, Park, Ye, Yang, Seo, Chang, and Seo]{chang2024largelanguagemodelsacquire}
Hoyeon Chang, Jinho Park, Seonghyeon Ye, Sohee Yang, Youngkyung Seo, Du-Seong Chang, and Minjoon Seo.
\newblock How do large language models acquire factual knowledge during pretraining?, 2024.
\newblock URL \url{https://arxiv.org/abs/2406.11813}.

\bibitem[Wu et~al.(2023)Wu, Jiang, Jiang, Xie, and Tu]{wu-etal-2023-plms}
Weiqi Wu, Chengyue Jiang, Yong Jiang, Pengjun Xie, and Kewei Tu.
\newblock Do {PLM}s know and understand ontological knowledge?
\newblock In Anna Rogers, Jordan Boyd-Graber, and Naoaki Okazaki, editors, \emph{Proceedings of the 61st Annual Meeting of the Association for Computational Linguistics (Volume 1: Long Papers)}, pages 3080--3101, Toronto, Canada, July 2023. Association for Computational Linguistics.
\newblock \doi{10.18653/v1/2023.acl-long.173}.
\newblock URL \url{https://aclanthology.org/2023.acl-long.173/}.

\bibitem[Petroni et~al.(2019)Petroni, Rockt{\"a}schel, Lewis, Bakhtin, Wu, Miller, and Riedel]{Petroni2019LanguageMA}
Fabio Petroni, Tim Rockt{\"a}schel, Patrick Lewis, Anton Bakhtin, Yuxiang Wu, Alexander~H. Miller, and Sebastian Riedel.
\newblock Language models as knowledge bases?
\newblock \emph{ArXiv}, abs/1909.01066, 2019.
\newblock URL \url{https://api.semanticscholar.org/CorpusID:202539551}.

\bibitem[Ding et~al.(2024)Ding, Fan, bo~Ning, Wang, Li, Yin, Chua, and Li]{Ding2024ASO}
Yujuan Ding, Wenqi Fan, Liang bo~Ning, Shijie Wang, Hengyun Li, Dawei Yin, Tat-Seng Chua, and Qing Li.
\newblock A survey on rag meets llms: Towards retrieval-augmented large language models.
\newblock \emph{ArXiv}, abs/2405.06211, 2024.
\newblock URL \url{https://api.semanticscholar.org/CorpusID:276185331}.

\bibitem[Lewis et~al.(2020)Lewis, Perez, Piktus, Petroni, Karpukhin, Goyal, Kuttler, Lewis, tau Yih, Rockt{\"a}schel, Riedel, and Kiela]{Lewis2020RetrievalAugmentedGF}
Patrick Lewis, Ethan Perez, Aleksandara Piktus, Fabio Petroni, Vladimir Karpukhin, Naman Goyal, Heinrich Kuttler, Mike Lewis, Wen tau Yih, Tim Rockt{\"a}schel, Sebastian Riedel, and Douwe Kiela.
\newblock Retrieval-augmented generation for knowledge-intensive nlp tasks.
\newblock \emph{ArXiv}, abs/2005.11401, 2020.
\newblock URL \url{https://api.semanticscholar.org/CorpusID:218869575}.

\bibitem[Ram et~al.(2023)Ram, Levine, Dalmedigos, Muhlgay, Shashua, Leyton-Brown, and Shoham]{ram2023incontextretrievalaugmentedlanguagemodels}
Ori Ram, Yoav Levine, Itay Dalmedigos, Dor Muhlgay, Amnon Shashua, Kevin Leyton-Brown, and Yoav Shoham.
\newblock In-context retrieval-augmented language models, 2023.
\newblock URL \url{https://arxiv.org/abs/2302.00083}.

\bibitem[Guu et~al.(2020)Guu, Lee, Tung, Pasupat, and Chang]{guu2020realm}
Kelvin Guu, Kenton Lee, Zora Tung, Panupong Pasupat, and Mingwei Chang.
\newblock Retrieval augmented language model pre-training.
\newblock In Hal~Daumé III and Aarti Singh, editors, \emph{Proceedings of the 37th International Conference on Machine Learning}, volume 119 of \emph{Proceedings of Machine Learning Research}, pages 3929--3938. PMLR, 13--18 Jul 2020.
\newblock URL \url{https://proceedings.mlr.press/v119/guu20a.html}.

\bibitem[Li et~al.(2025{\natexlab{a}})Li, Dong, Jin, Zhang, Zhou, Zhu, Zhang, and Dou]{li2025searcho1agenticsearchenhancedlarge}
Xiaoxi Li, Guanting Dong, Jiajie Jin, Yuyao Zhang, Yujia Zhou, Yutao Zhu, Peitian Zhang, and Zhicheng Dou.
\newblock Search-o1: Agentic search-enhanced large reasoning models, 2025{\natexlab{a}}.
\newblock URL \url{https://arxiv.org/abs/2501.05366}.

\bibitem[Jin et~al.(2025)Jin, Zeng, Yue, Yoon, Arik, Wang, Zamani, and Han]{jin2025searchr1trainingllmsreason}
Bowen Jin, Hansi Zeng, Zhenrui Yue, Jinsung Yoon, Sercan Arik, Dong Wang, Hamed Zamani, and Jiawei Han.
\newblock Search-r1: Training llms to reason and leverage search engines with reinforcement learning, 2025.
\newblock URL \url{https://arxiv.org/abs/2503.09516}.

\bibitem[Devlin et~al.(2019)Devlin, Chang, Lee, and Toutanova]{devlin-etal-2019-bert}
Jacob Devlin, Ming-Wei Chang, Kenton Lee, and Kristina Toutanova.
\newblock {BERT}: Pre-training of deep bidirectional transformers for language understanding.
\newblock In Jill Burstein, Christy Doran, and Thamar Solorio, editors, \emph{Proceedings of the 2019 Conference of the North {A}merican Chapter of the Association for Computational Linguistics: Human Language Technologies, Volume 1 (Long and Short Papers)}, pages 4171--4186, Minneapolis, Minnesota, June 2019. Association for Computational Linguistics.
\newblock \doi{10.18653/v1/N19-1423}.
\newblock URL \url{https://aclanthology.org/N19-1423/}.

\bibitem[Yu et~al.(2025)Yu, Zhang, Zhu, Yuan, Zuo, Yue, Fan, Liu, Liu, Liu, et~al.]{yu2025dapoopensourcellmreinforcement}
Qiying Yu, Zheng Zhang, Ruofei Zhu, Yufeng Yuan, Xiaochen Zuo, Yu~Yue, Tiantian Fan, Gaohong Liu, Lingjun Liu, Xin Liu, et~al.
\newblock Dapo: An open-source llm reinforcement learning system at scale, 2025.
\newblock URL \url{https://arxiv.org/abs/2503.14476}.

\bibitem[Yan et~al.(2024)Yan, Gu, Zhu, and Ling]{Yan2024CorrectiveRA}
Shi-Qi Yan, Jia-Chen Gu, Yun Zhu, and Zhen-Hua Ling.
\newblock Corrective retrieval augmented generation.
\newblock \emph{ArXiv}, abs/2401.15884, 2024.
\newblock URL \url{https://api.semanticscholar.org/CorpusID:267312595}.

\bibitem[Gao et~al.(2023)Gao, Xiong, Gao, Jia, Pan, Bi, Dai, Sun, Guo, Wang, and Wang]{Gao2023RetrievalAugmentedGF}
Yunfan Gao, Yun Xiong, Xinyu Gao, Kangxiang Jia, Jinliu Pan, Yuxi Bi, Yi~Dai, Jiawei Sun, Qianyu Guo, Meng Wang, and Haofen Wang.
\newblock Retrieval-augmented generation for large language models: A survey.
\newblock \emph{ArXiv}, abs/2312.10997, 2023.
\newblock URL \url{https://api.semanticscholar.org/CorpusID:266359151}.

\bibitem[Zhang et~al.(2024{\natexlab{a}})Zhang, Patil, Jain, Shen, Zaharia, Stoica, and Gonzalez]{Zhang2024RAFTAL}
Tianjun Zhang, Shishir~G. Patil, Naman Jain, Sheng Shen, Matei~A. Zaharia, Ion Stoica, and Joseph Gonzalez.
\newblock Raft: Adapting language model to domain specific rag.
\newblock \emph{ArXiv}, abs/2403.10131, 2024{\natexlab{a}}.
\newblock URL \url{https://api.semanticscholar.org/CorpusID:268510197}.

\bibitem[Gilbert et~al.(2024)Gilbert, Kather, and Hogan]{Gilbert2024AugmentedNL}
Stephen Gilbert, Jakob~Nikolas Kather, and Aidan Hogan.
\newblock Augmented non-hallucinating large language models as medical information curators.
\newblock \emph{NPJ Digital Medicine}, 7, 2024.
\newblock URL \url{https://api.semanticscholar.org/CorpusID:269327466}.

\bibitem[Li et~al.(2024)Li, Yuan, and Zhang]{li2024enhancingllmfactualaccuracy}
Jiarui Li, Ye~Yuan, and Zehua Zhang.
\newblock Enhancing llm factual accuracy with rag to counter hallucinations: A case study on domain-specific queries in private knowledge-bases, 2024.
\newblock URL \url{https://arxiv.org/abs/2403.10446}.

\bibitem[Huang et~al.(2023)Huang, Yu, Ma, Zhong, Feng, Wang, Chen, Peng, Feng, Qin, and Liu]{Huang2023ASO}
Lei Huang, Weijiang Yu, Weitao Ma, Weihong Zhong, Zhangyin Feng, Haotian Wang, Qianglong Chen, Weihua Peng, Xiaocheng Feng, Bing Qin, and Ting Liu.
\newblock A survey on hallucination in large language models: Principles, taxonomy, challenges, and open questions.
\newblock \emph{ArXiv}, abs/2311.05232, 2023.
\newblock URL \url{https://api.semanticscholar.org/CorpusID:265067168}.

\bibitem[Singh et~al.(2025)Singh, Ehtesham, Kumar, and Khoei]{singh2025agenticretrievalaugmentedgenerationsurvey}
Aditi Singh, Abul Ehtesham, Saket Kumar, and Tala~Talaei Khoei.
\newblock Agentic retrieval-augmented generation: A survey on agentic rag, 2025.
\newblock URL \url{https://arxiv.org/abs/2501.09136}.

\bibitem[Ravuru et~al.(2024)Ravuru, Sakhinana, and Runkana]{ravuru2024agenticretrievalaugmentedgenerationtime}
Chidaksh Ravuru, Sagar~Srinivas Sakhinana, and Venkataramana Runkana.
\newblock Agentic retrieval-augmented generation for time series analysis, 2024.
\newblock URL \url{https://arxiv.org/abs/2408.14484}.

\bibitem[An et~al.(2024)An, Ding, Fu, Chu, Li, and Du]{an2024goldenretrieverhighfidelityagenticretrieval}
Zhiyu An, Xianzhong Ding, Yen-Chun Fu, Cheng-Chung Chu, Yan Li, and Wan Du.
\newblock Golden-retriever: High-fidelity agentic retrieval augmented generation for industrial knowledge base, 2024.
\newblock URL \url{https://arxiv.org/abs/2408.00798}.

\bibitem[Wang et~al.(2025)Wang, Ding, Chen, Wu, Wang, Xie, and Zhao]{wang2025vidoragvisualdocumentretrievalaugmented}
Qiuchen Wang, Ruixue Ding, Zehui Chen, Weiqi Wu, Shihang Wang, Pengjun Xie, and Feng Zhao.
\newblock Vidorag: Visual document retrieval-augmented generation via dynamic iterative reasoning agents, 2025.
\newblock URL \url{https://arxiv.org/abs/2502.18017}.

\bibitem[He et~al.(2025)He, Huang, Feng, Lin, Zhang, Li, and E]{he2025pasallmagentcomprehensive}
Yichen He, Guanhua Huang, Peiyuan Feng, Yuan Lin, Yuchen Zhang, Hang Li, and Weinan E.
\newblock Pasa: An llm agent for comprehensive academic paper search, 2025.
\newblock URL \url{https://arxiv.org/abs/2501.10120}.

\bibitem[Alzubi et~al.(2025)Alzubi, Brooks, Chiniya, Contente, von Gerlach, Irwin, Jiang, Kaz, Nguyen, Oh, Tyagi, and Viswanath]{alzubi2025opendeepsearchdemocratizing}
Salaheddin Alzubi, Creston Brooks, Purva Chiniya, Edoardo Contente, Chiara von Gerlach, Lucas Irwin, Yihan Jiang, Arda Kaz, Windsor Nguyen, Sewoong Oh, Himanshu Tyagi, and Pramod Viswanath.
\newblock Open deep search: Democratizing search with open-source reasoning agents, 2025.
\newblock URL \url{https://arxiv.org/abs/2503.20201}.

\bibitem[Gur et~al.(2023)Gur, Furuta, Huang, Safdari, Matsuo, Eck, and Faust]{Gur2023ARW}
Izzeddin Gur, Hiroki Furuta, Austin Huang, Mustafa Safdari, Yutaka Matsuo, Douglas Eck, and Aleksandra Faust.
\newblock A real-world webagent with planning, long context understanding, and program synthesis.
\newblock \emph{ArXiv}, abs/2307.12856, 2023.
\newblock URL \url{https://api.semanticscholar.org/CorpusID:260126067}.

\bibitem[Zhang et~al.(2024{\natexlab{b}})Zhang, Li, Li, Shi, and Jin]{zhang2024codeagent}
Kechi Zhang, Jia Li, Ge~Li, Xianjie Shi, and Zhi Jin.
\newblock Codeagent: Enhancing code generation with tool-integrated agent systems for real-world repo-level coding challenges, 2024{\natexlab{b}}.

\bibitem[Gundawar et~al.(2024)Gundawar, Verma, Guan, Valmeekam, Bhambri, and Kambhampati]{gundawar2024robustplanningllmmoduloframework}
Atharva Gundawar, Mudit Verma, Lin Guan, Karthik Valmeekam, Siddhant Bhambri, and Subbarao Kambhampati.
\newblock Robust planning with llm-modulo framework: Case study in travel planning, 2024.
\newblock URL \url{https://arxiv.org/abs/2405.20625}.

\bibitem[Shinn et~al.(2024)Shinn, Cassano, Gopinath, Narasimhan, and Yao]{shinn2024reflexion}
Noah Shinn, Federico Cassano, Ashwin Gopinath, Karthik Narasimhan, and Shunyu Yao.
\newblock Reflexion: Language agents with verbal reinforcement learning.
\newblock \emph{Advances in Neural Information Processing Systems}, 36, 2024.

\bibitem[Park et~al.(2023)Park, O'Brien, Cai, Morris, Liang, and Bernstein]{park2023generativeagentsinteractivesimulacra}
Joon~Sung Park, Joseph~C. O'Brien, Carrie~J. Cai, Meredith~Ringel Morris, Percy Liang, and Michael~S. Bernstein.
\newblock Generative agents: Interactive simulacra of human behavior, 2023.
\newblock URL \url{https://arxiv.org/abs/2304.03442}.

\bibitem[Yao et~al.(2023{\natexlab{a}})Yao, Zhao, Yu, Du, Shafran, Narasimhan, and Cao]{yao2023react}
Shunyu Yao, Jeffrey Zhao, Dian Yu, Nan Du, Izhak Shafran, Karthik Narasimhan, and Yuan Cao.
\newblock {ReAct}: Synergizing reasoning and acting in language models.
\newblock In \emph{International Conference on Learning Representations (ICLR)}, 2023{\natexlab{a}}.

\bibitem[Jiabin~Tang(2025)]{AutoAgent}
Chao~Huang Jiabin~Tang, Tianyu~Fan.
\newblock {AutoAgent: A Fully-Automated and Zero-Code Framework for LLM Agents}, 2025.
\newblock URL \url{https://arxiv.org/abs/2502.05957}.

\bibitem[Putta et~al.(2024)Putta, Mills, Garg, Motwani, Finn, Garg, and Rafailov]{putta2024agent}
Pranav Putta, Edmund Mills, Naman Garg, Sumeet Motwani, Chelsea Finn, Divyansh Garg, and Rafael Rafailov.
\newblock Agent q: Advanced reasoning and learning for autonomous ai agents.
\newblock \emph{arXiv preprint arXiv:2408.07199}, 2024.

\bibitem[Feng et~al.(2024)Feng, He, Huang, Lin, Zhang, Zhang, and Li]{feng2024agile}
Peiyuan Feng, Yichen He, Guanhua Huang, Yuan Lin, Hanchong Zhang, Yuchen Zhang, and Hang Li.
\newblock Agile: A novel framework of llm agents.
\newblock \emph{arXiv preprint arXiv:2405.14751}, 2024.

\bibitem[Li et~al.(2025{\natexlab{b}})Li, Xue, Zhang, Yang, Zhang, Wang, Yu, Hui, Lin, and Liu]{li2025startselftaughtreasonertools}
Chengpeng Li, Mingfeng Xue, Zhenru Zhang, Jiaxi Yang, Beichen Zhang, Xiang Wang, Bowen Yu, Binyuan Hui, Junyang Lin, and Dayiheng Liu.
\newblock Start: Self-taught reasoner with tools, 2025{\natexlab{b}}.
\newblock URL \url{https://arxiv.org/abs/2503.04625}.

\bibitem[Jaech et~al.(2024)Jaech, Kalai, Lerer, Richardson, El-Kishky, Low, Helyar, Madry, Beutel, Carney, et~al.]{openai2024openaio1card}
Aaron Jaech, Adam Kalai, Adam Lerer, Adam Richardson, Ahmed El-Kishky, Aiden Low, Alec Helyar, Aleksander Madry, Alex Beutel, Alex Carney, et~al.
\newblock Openai o1 system card, 2024.
\newblock URL \url{https://arxiv.org/abs/2412.16720}.

\bibitem[Muennighoff et~al.(2025)Muennighoff, Yang, Shi, Li, Fei-Fei, Hajishirzi, Zettlemoyer, Liang, Candès, and Hashimoto]{muennighoff2025s1simpletesttimescaling}
Niklas Muennighoff, Zitong Yang, Weijia Shi, Xiang~Lisa Li, Li~Fei-Fei, Hannaneh Hajishirzi, Luke Zettlemoyer, Percy Liang, Emmanuel Candès, and Tatsunori Hashimoto.
\newblock s1: Simple test-time scaling, 2025.
\newblock URL \url{https://arxiv.org/abs/2501.19393}.

\bibitem[Snell et~al.(2024)Snell, Lee, Xu, and Kumar]{Snell2024ScalingLT}
Charlie Snell, Jaehoon Lee, Kelvin Xu, and Aviral Kumar.
\newblock Scaling llm test-time compute optimally can be more effective than scaling model parameters.
\newblock \emph{ArXiv}, abs/2408.03314, 2024.
\newblock URL \url{https://api.semanticscholar.org/CorpusID:271719990}.

\bibitem[Wei et~al.(2022)Wei, Wang, Schuurmans, Bosma, ichter, Xia, Chi, Le, and Zhou]{Wu2023ChainOT}
Jason Wei, Xuezhi Wang, Dale Schuurmans, Maarten Bosma, brian ichter, Fei Xia, Ed~Chi, Quoc~V Le, and Denny Zhou.
\newblock Chain-of-thought prompting elicits reasoning in large language models.
\newblock In S.~Koyejo, S.~Mohamed, A.~Agarwal, D.~Belgrave, K.~Cho, and A.~Oh, editors, \emph{Advances in Neural Information Processing Systems}, volume~35, pages 24824--24837. Curran Associates, Inc., 2022.
\newblock URL \url{https://proceedings.neurips.cc/paper_files/paper/2022/file/9d5609613524ecf4f15af0f7b31abca4-Paper-Conference.pdf}.

\bibitem[Yao et~al.(2023{\natexlab{b}})Yao, Yu, Zhao, Shafran, Griffiths, Cao, and Narasimhan]{yao2023treethoughtsdeliberateproblem}
Shunyu Yao, Dian Yu, Jeffrey Zhao, Izhak Shafran, Thomas~L. Griffiths, Yuan Cao, and Karthik Narasimhan.
\newblock Tree of thoughts: Deliberate problem solving with large language models, 2023{\natexlab{b}}.
\newblock URL \url{https://arxiv.org/abs/2305.10601}.

\bibitem[Yugeswardeenoo et~al.(2024)Yugeswardeenoo, Zhu, and O'Brien]{yugeswardeenoo2024questionanalysispromptingimprovesllm}
Dharunish Yugeswardeenoo, Kevin Zhu, and Sean O'Brien.
\newblock Question-analysis prompting improves llm performance in reasoning tasks, 2024.
\newblock URL \url{https://arxiv.org/abs/2407.03624}.

\bibitem[Fu et~al.(2023{\natexlab{a}})Fu, Peng, Sabharwal, Clark, and Khot]{fu2023complexitybasedpromptingmultistepreasoning}
Yao Fu, Hao Peng, Ashish Sabharwal, Peter Clark, and Tushar Khot.
\newblock Complexity-based prompting for multi-step reasoning, 2023{\natexlab{a}}.
\newblock URL \url{https://arxiv.org/abs/2210.00720}.

\bibitem[Subramaniam et~al.(2025)Subramaniam, Du, Tenenbaum, Torralba, Li, and Mordatch]{subramaniam2025multiagentfinetuningselfimprovement}
Vighnesh Subramaniam, Yilun Du, Joshua~B. Tenenbaum, Antonio Torralba, Shuang Li, and Igor Mordatch.
\newblock Multiagent finetuning: Self improvement with diverse reasoning chains, 2025.
\newblock URL \url{https://arxiv.org/abs/2501.05707}.

\bibitem[Lin et~al.(2024)Lin, Chen, Chen, Shi, Lomeli, James, Rodriguez, Kahn, Szilvasy, Lewis, Zettlemoyer, and Yih]{lin2024raditretrievalaugmenteddualinstruction}
Xi~Victoria Lin, Xilun Chen, Mingda Chen, Weijia Shi, Maria Lomeli, Rich James, Pedro Rodriguez, Jacob Kahn, Gergely Szilvasy, Mike Lewis, Luke Zettlemoyer, and Scott Yih.
\newblock Ra-dit: Retrieval-augmented dual instruction tuning, 2024.
\newblock URL \url{https://arxiv.org/abs/2310.01352}.

\bibitem[Mukherjee et~al.(2023)Mukherjee, Mitra, Jawahar, Agarwal, Palangi, and Awadallah]{mukherjee2023orcaprogressivelearningcomplex}
Subhabrata Mukherjee, Arindam Mitra, Ganesh Jawahar, Sahaj Agarwal, Hamid Palangi, and Ahmed Awadallah.
\newblock Orca: Progressive learning from complex explanation traces of gpt-4, 2023.
\newblock URL \url{https://arxiv.org/abs/2306.02707}.

\bibitem[Srivastava et~al.(2025)Srivastava, Cao, and Wang]{srivastava2025reasoningabilitysmalllanguage}
Gaurav Srivastava, Shuxiang Cao, and Xuan Wang.
\newblock Towards reasoning ability of small language models, 2025.
\newblock URL \url{https://arxiv.org/abs/2502.11569}.

\bibitem[Shridhar et~al.(2023)Shridhar, Stolfo, and Sachan]{shridhar-etal-2023-distilling}
Kumar Shridhar, Alessandro Stolfo, and Mrinmaya Sachan.
\newblock Distilling reasoning capabilities into smaller language models.
\newblock In Anna Rogers, Jordan Boyd-Graber, and Naoaki Okazaki, editors, \emph{Findings of the Association for Computational Linguistics: ACL 2023}, pages 7059--7073, Toronto, Canada, July 2023. Association for Computational Linguistics.
\newblock \doi{10.18653/v1/2023.findings-acl.441}.
\newblock URL \url{https://aclanthology.org/2023.findings-acl.441/}.

\bibitem[Fu et~al.(2023{\natexlab{b}})Fu, Peng, Ou, Sabharwal, and Khot]{fu2023specializingsmallerlanguagemodels}
Yao Fu, Hao Peng, Litu Ou, Ashish Sabharwal, and Tushar Khot.
\newblock Specializing smaller language models towards multi-step reasoning, 2023{\natexlab{b}}.
\newblock URL \url{https://arxiv.org/abs/2301.12726}.

\bibitem[Cole et~al.(2023)Cole, Chaudhary, Dhingra, and Talukdar]{cole2023salientspanmaskingtemporal}
Jeremy~R. Cole, Aditi Chaudhary, Bhuwan Dhingra, and Partha Talukdar.
\newblock Salient span masking for temporal understanding, 2023.
\newblock URL \url{https://arxiv.org/abs/2303.12860}.

\bibitem[Yang et~al.(2018)Yang, Qi, Zhang, Bengio, Cohen, Salakhutdinov, and Manning]{yang2018hotpotqa}
Zhilin Yang, Peng Qi, Saizheng Zhang, Yoshua Bengio, William~W. Cohen, Ruslan Salakhutdinov, and Christopher~D. Manning.
\newblock {HotpotQA}: A dataset for diverse, explainable multi-hop question answering.
\newblock In \emph{Conference on Empirical Methods in Natural Language Processing ({EMNLP})}, 2018.

\bibitem[Zhu et~al.(2024)Zhu, Hwang, Dugan, and Callison-Burch]{zhu-etal-2024-fanoutqa}
Andrew Zhu, Alyssa Hwang, Liam Dugan, and Chris Callison-Burch.
\newblock {F}an{O}ut{QA}: A multi-hop, multi-document question answering benchmark for large language models.
\newblock In Lun-Wei Ku, Andre Martins, and Vivek Srikumar, editors, \emph{Proceedings of the 62nd Annual Meeting of the Association for Computational Linguistics (Volume 2: Short Papers)}, pages 18--37, Bangkok, Thailand, August 2024. Association for Computational Linguistics.
\newblock \doi{10.18653/v1/2024.acl-short.2}.
\newblock URL \url{https://aclanthology.org/2024.acl-short.2/}.

\bibitem[Trivedi et~al.(2022)Trivedi, Balasubramanian, Khot, and Sabharwal]{trivedi2021musique}
Harsh Trivedi, Niranjan Balasubramanian, Tushar Khot, and Ashish Sabharwal.
\newblock {M}u{S}i{Q}ue: Multihop questions via single-hop question composition.
\newblock \emph{Transactions of the Association for Computational Linguistics}, 2022.

\bibitem[Ho et~al.(2020)Ho, Duong~Nguyen, Sugawara, and Aizawa]{xanh2020_2wikimultihop}
Xanh Ho, Anh-Khoa Duong~Nguyen, Saku Sugawara, and Akiko Aizawa.
\newblock Constructing a multi-hop {QA} dataset for comprehensive evaluation of reasoning steps.
\newblock In \emph{Proceedings of the 28th International Conference on Computational Linguistics}, pages 6609--6625, Barcelona, Spain (Online), December 2020. International Committee on Computational Linguistics.
\newblock URL \url{https://www.aclweb.org/anthology/2020.coling-main.580}.

\bibitem[Press et~al.(2023)Press, Zhang, Min, Schmidt, Smith, and Lewis]{press-etal-2023-measuring}
Ofir Press, Muru Zhang, Sewon Min, Ludwig Schmidt, Noah Smith, and Mike Lewis.
\newblock Measuring and narrowing the compositionality gap in language models.
\newblock In Houda Bouamor, Juan Pino, and Kalika Bali, editors, \emph{Findings of the Association for Computational Linguistics: EMNLP 2023}, pages 5687--5711, Singapore, December 2023. Association for Computational Linguistics.
\newblock \doi{10.18653/v1/2023.findings-emnlp.378}.
\newblock URL \url{https://aclanthology.org/2023.findings-emnlp.378/}.

\bibitem[Vu et~al.(2023)Vu, Iyyer, Wang, Constant, Wei, Wei, Tar, Sung, Zhou, Le, and Luong]{vu2023freshllmsrefreshinglargelanguage}
Tu~Vu, Mohit Iyyer, Xuezhi Wang, Noah Constant, Jerry Wei, Jason Wei, Chris Tar, Yun-Hsuan Sung, Denny Zhou, Quoc Le, and Thang Luong.
\newblock Freshllms: Refreshing large language models with search engine augmentation, 2023.
\newblock URL \url{https://arxiv.org/abs/2310.03214}.

\bibitem[Khandelwal et~al.(2020)Khandelwal, Levy, Jurafsky, Zettlemoyer, and Lewis]{khandelwal2020generalizationmemorizationnearestneighbor}
Urvashi Khandelwal, Omer Levy, Dan Jurafsky, Luke Zettlemoyer, and Mike Lewis.
\newblock Generalization through memorization: Nearest neighbor language models, 2020.
\newblock URL \url{https://arxiv.org/abs/1911.00172}.

\bibitem[Borgeaud et~al.(2022)Borgeaud, Mensch, Hoffmann, Cai, Rutherford, Millican, van~den Driessche, Lespiau, Damoc, Clark, de~Las~Casas, Guy, Menick, Ring, Hennigan, Huang, Maggiore, Jones, Cassirer, Brock, Paganini, Irving, Vinyals, Osindero, Simonyan, Rae, Elsen, and Sifre]{borgeaud2022improvinglanguagemodelsretrieving}
Sebastian Borgeaud, Arthur Mensch, Jordan Hoffmann, Trevor Cai, Eliza Rutherford, Katie Millican, George van~den Driessche, Jean-Baptiste Lespiau, Bogdan Damoc, Aidan Clark, Diego de~Las~Casas, Aurelia Guy, Jacob Menick, Roman Ring, Tom Hennigan, Saffron Huang, Loren Maggiore, Chris Jones, Albin Cassirer, Andy Brock, Michela Paganini, Geoffrey Irving, Oriol Vinyals, Simon Osindero, Karen Simonyan, Jack~W. Rae, Erich Elsen, and Laurent Sifre.
\newblock Improving language models by retrieving from trillions of tokens, 2022.
\newblock URL \url{https://arxiv.org/abs/2112.04426}.

\bibitem[Izacard et~al.(2022)Izacard, Lewis, Lomeli, Hosseini, Petroni, Schick, Dwivedi-Yu, Joulin, Riedel, and Grave]{izacard2022atlasfewshotlearningretrieval}
Gautier Izacard, Patrick Lewis, Maria Lomeli, Lucas Hosseini, Fabio Petroni, Timo Schick, Jane Dwivedi-Yu, Armand Joulin, Sebastian Riedel, and Edouard Grave.
\newblock Atlas: Few-shot learning with retrieval augmented language models, 2022.
\newblock URL \url{https://arxiv.org/abs/2208.03299}.

\bibitem[Shao et~al.(2024)Shao, Wang, Zhu, Xu, Song, Zhang, Li, Wu, and Guo]{DBLP:deepseekmath}
Zhihong Shao, Peiyi Wang, Qihao Zhu, Runxin Xu, Junxiao Song, Mingchuan Zhang, Y.~K. Li, Y.~Wu, and Daya Guo.
\newblock Deepseekmath: Pushing the limits of mathematical reasoning in open language models.
\newblock \emph{CoRR}, abs/2402.03300, 2024.
\newblock \doi{10.48550/ARXIV.2402.03300}.
\newblock URL \url{https://doi.org/10.48550/arXiv.2402.03300}.

\end{thebibliography}
